%% file: isac.tex
\documentclass[letterpaper, 10 pt, journal]{IEEEtran}

\IEEEoverridecommandlockouts                              
                                                          
\usepackage{graphicx} 
\usepackage[flushleft]{threeparttable}
\usepackage{tabularx}
\usepackage{multirow}
\usepackage{amsmath} 
\usepackage{amssymb}  
\usepackage{color}
\usepackage[noadjust]{cite}
\usepackage{url}
\usepackage[final]{changes}

\newtheorem{Lemma}{Lemma}
\newtheorem{Assumption}{Assumption}
\newtheorem{Theorem}{Theorem}

\newtheorem{Remark}{Remark}
\newtheorem{Proposition}{Proposition}
\newtheorem{Definition}{Definition}

\title{Iterative Sequential Action Control for Stable, Model-Based Control of Nonlinear Systems}

\author{Emmanouil Tzorakoleftherakis, \IEEEmembership{Member, ̃IEEE}, 
and Todd D. Murphey, \IEEEmembership{Member, ̃IEEE}
\thanks{Authors are with the Neuroscience and Robotics Laboratory (N$\times$R) at the Department of Mechanical Engineering, Northwestern University, Evanston, IL.
\tt\scriptsize Email: man7therakis@u.northwestern.edu, t-murphey@northwestern.edu.}
}

\begin{document}

\maketitle
\thispagestyle{empty}
\pagestyle{empty}

\begin{abstract}
This paper presents iterative Sequential Action Control (iSAC), a receding horizon approach for control of nonlinear systems. The iSAC method has a closed-form open-loop solution, which is iteratively updated between time steps by introducing constant control values applied for short duration. Application of a contractive constraint on the cost is shown to lead to closed-loop asymptotic stability under mild assumptions. The effect of asymptotically decaying disturbances on system trajectories is also examined. To demonstrate the applicability of iSAC, we employ a variety of systems and conditions, including a 13-dimensional quaternion-based quadrotor and NASA's TRACE spacecraft. Each system is tested in different scenarios, ranging from feasible and infeasible trajectory tracking, to setpoint stabilization, with or without the presence of external disturbances. Finally, limitations of this work are discussed.

\end{abstract}

\section{Introduction}
Model-predictive control (MPC) has been widely used over the past twenty years for control of linear and nonlinear systems \cite{mayne2000constrained,mayne2014model}. The rationale behind MPC is the following: at each time step, an $L_2$ or alternative variation of the cost function is locally optimized over time to obtain the open-loop control as a function of time, only a small portion of which is actually applied to the system. The time horizon is then shifted, and the process is repeated based on acquired state feedback. Using this approach, it is clear that optimizing with respect to the entire open-loop control, although often effective, might not be the most efficient way to compute the control since most of the optimizer is typically discarded. 

In view of this observation, our previous work in \cite{SACtro,SACexosekApplic,tzorakoleftherakis2016model,Fan-RSS-16,mamakoukas2016sequential} presented Sequential Action Control (SAC), a receding horizon approach for control of nonlinear systems, that exploits elements from hybrid systems theory \cite{eger,egerstedt2006transition}. In contrast with the aforementioned MPC methods, the open-loop solution in SAC optimizes the needle variation \cite{pontryagin1987mathematical} of the cost resulting in a single, constant magnitude action which does not optimize, but rather improves the cost function value relative to applying only a nominal control signal. Since only a single action---obtained in closed-form---is computed at each time step, control calculation is efficient. Our earlier work in \cite{SACtro,SACexosekApplic,tzorakoleftherakis2016model,Fan-RSS-16,mamakoukas2016sequential} indicates that SAC can drive benchmark and challenging systems---including the cart-pendulum, acrobot and pendubot in \cite{SACtro}, hopping and humanoid locomotion in \cite{tzorakoleftherakis2016model} and rotor vehicles with dynamics on Lie groups in \cite{Fan-RSS-16}---close to a desired equilibrium. \added{Nevertheless, it cannot achieve final stabilization and, as a result, switching to a locally stabilizing controller is necessary \cite{SACtro,Fan-RSS-16}. An illustration of this behavior is shown in Fig.~\ref{fig:sac_cartpend} for the cart-pendulum inversion system.}

\added{This paper presents iterative Sequential Action Control (iSAC), an extension of SAC that addresses one fundamental question that was left open in our previous work---how can SAC achieve consistent stabilizing behavior? We provide new theoretical results that show that our modified method, iSAC, can achieve closed-loop stability, which was not possible using SAC in our previous work. We also present side-by-side the procedural differences between iSAC and SAC and how the modifications allow us to prove closed-loop stability.
Simulation results show that, with minimal modifications across examples, iSAC is consistently successful in a variety of control scenarios, ranging from benchmark inversion problems \cite{aastrom2000swinging} to control of quadrotors performing complex three-dimensional tasks like flips, and constrained maneuvers of spacecrafts. Some of the examined conditions include feasible and infeasible trajectory tracking, setpoint stabilization, and control under external disturbances.}

In addition to the aforementioned points, the following novelties of iSAC distinguish this work from alternative MPC methods (see, e.g.,~\cite{lizarralde1997feedback,da2008interactive,tassa2008receding,tassa2012synthesis,chen1998nonlinear,mayne2000constrained,grune2011nonlinear,mayne2014model} and references therein). In order to solve the open-loop problem, most MPC methods either employ nonlinear programming solvers (see \cite{wachter2006implementation} and \cite{biegler2013survey} for a review) or solve a matrix Riccati differential equation as, for example, in \cite{da2008interactive,tassa2008receding,tassa2012synthesis}. On the contrary, the solution of the open-loop problem in iSAC has an analytic form which requires only forward simulations of the dynamics and a costate variable, which is computationally efficient and does not depend on black-box optimization routines. Moreover, control saturations can be incorporated without additional computational overhead. Finally, as opposed to many MPC alternatives that utilize discrete-time dynamics \cite{mayne2000constrained,chen1997quasi,chen1998nonlinear,tassa2012synthesis,mayne2014model}, iSAC uses continuous-time dynamics. As a result, variable-step integration may be used, which, combined with the previous points, significantly speeds-up the solution process.

\deleted{To achieve closed-loop stability for iSAC, we apply a contractive constraint on the cost \cite{de2000contractive,camponogara2002distributed,xie2008first,ferrari2009model}. Contractive constraints have been widely used in the MPC literature to show closed-loop stability as an alternative to methods relying on a terminal (region) constraint \cite{chen1998nonlinear,mayne2000constrained,fontes2001general,grune2011nonlinear,lee2011model,mayne2014model}. A disadvantage of these methods is that they require the computation of a terminal region, which has to be calculated separately for each system of interest. Many procedures have been proposed on how to find this region; see, e.g.,  \cite{chen1997quasi,fontes2001general,chen2003terminal,yu2009enlarging,SOSderivParrilo2000} and references therein. On the contrary, the contractive constraint approach can be applied to a variety of examples without modification and is easier to implement. In our case, a contractive constraint on the cost can be naturally integrated in iSAC, without additional work, by exploiting tools from hybrid systems theory, e.g., the mode insertion gradient (see \cite{eger,egerstedt2006transition}), as will be explained in Section~\ref{isac}.}

This paper is structured as follows: Section~\ref{prelims} provides a description of our previous work on SAC; Section~\ref{isac} describes iSAC and the features that differentiate the proposed method from SAC. In Section~\ref{stability} we discuss the global stability properties of iSAC. Section~\ref{simulationresults} demonstrates applicability of iSAC to a variety of systems and control scenarios while conclusive remarks are given in Section~\ref{conclusion}. Finally, all proofs are presented in the Appendix.


\section{Preliminaries - Sequential Action Control}
\input{sac}


\section{From SAC to {i}SAC}
\input{isac_desc}


\deleted{Local Stability Analysis section and proofs}

\section{Global Stability Analysis}
\input{stability}



\section{Simulation Results}
\input{sim_results}


\section{Conclusion}
\input{conclusion}

\appendix
\input{appendix}





\section*{ACKNOWLEDGMENT}

This material is based upon work supported by the National Science Foundation under awards CNS-1426961 and DCSD-1662233. Any opinions, findings, and conclusions or recommendations expressed in this material are those of the authors and do not necessarily reflect the views of the National Science Foundation.


\bibliographystyle{IEEEtran}
\bibliography{IEEEabrv,isac_journal}

\begin{IEEEbiography}[{\includegraphics[width=1in,height=1.25in,clip,keepaspectratio]{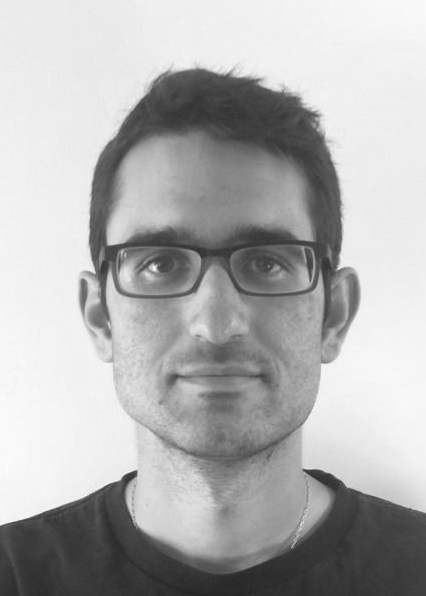}}]{Emmanouil Tzorakoleftherakis}
received the joint B.S./M.S. degree
in electrical and computer engineering from University of Patras, Greece in 2012, and the M.S. and Ph.D. degrees in mechanical
engineering from Northwestern University, Evanston, IL,
USA, in 2015 and 2017 respectively. His interests include control and robotics. 
\end{IEEEbiography}

\begin{IEEEbiography}[{\includegraphics[width=1in,height=1.25in,clip,keepaspectratio]{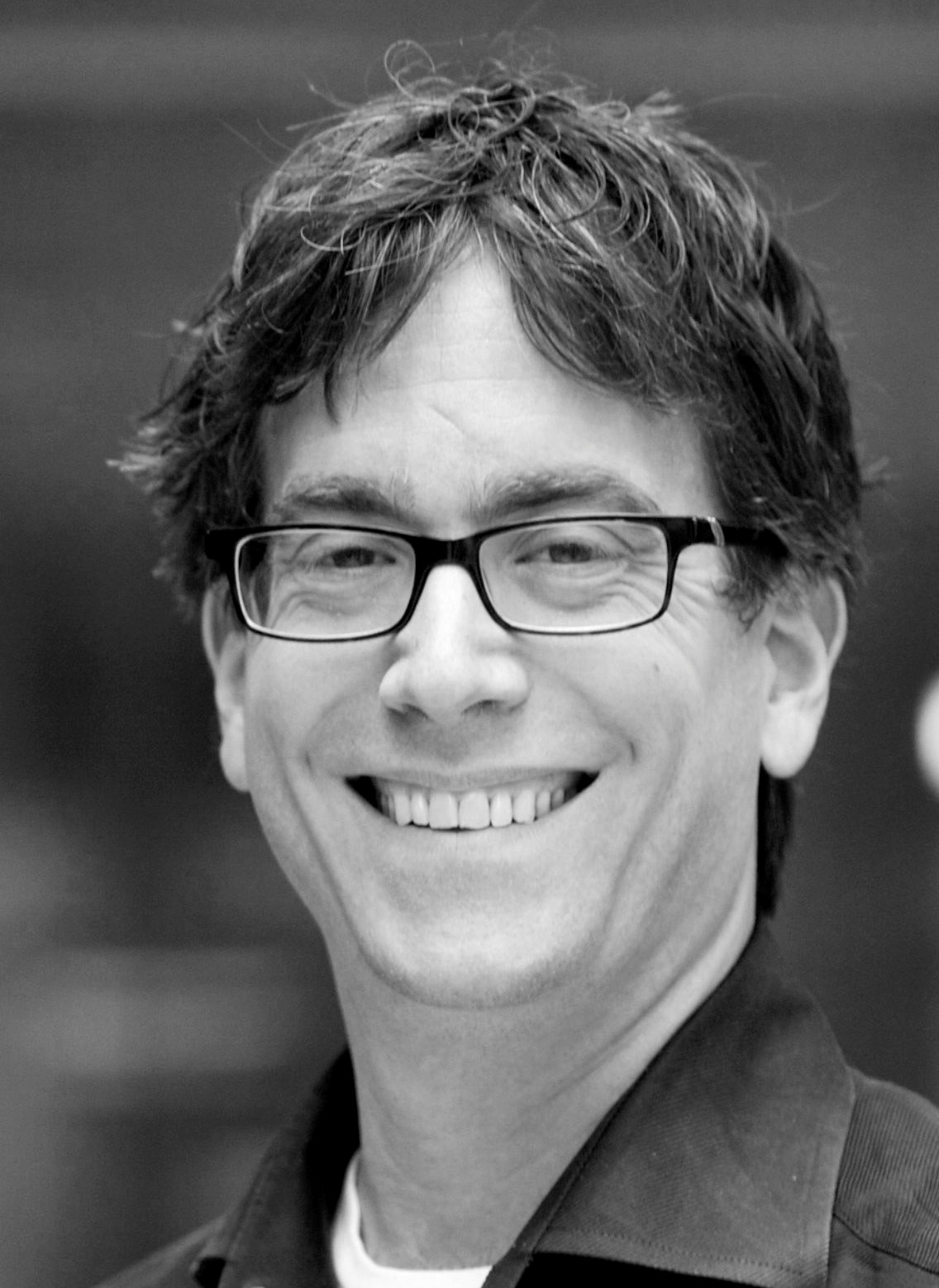}}]{Todd D. Murphey}
received the B.S. degree in mathematics  from  University  of  Arizona,  Tucson,  AZ,
USA,  and  the  Ph.D.  degree  in  control  and  dynamical systems from California Institute of Technology,
Pasadena, CA, USA. 

He is an Associate Professor of Mechanical Engineering with Northwestern University Evanston, IL, USA. His laboratory is part of the Neuroscience and Robotics  Laboratory,  and  his  research  interests  include  robotics,  control, and computational  methods  for biomechanical  systems and neuroscience. He  has  received  honors  including  the  National  Science  Foundation CAREER  award  in  2006,  membership  in  the  2014-2015  DARPA/IDA  Defense  Science  Study  Group,  and  Northwestern's  Professorship  of  Teaching Excellence. 
\end{IEEEbiography}

\end{document}

%% file: sac.tex
\label{prelims}

\begin{figure}
\centering
\includegraphics[width=3.3in]{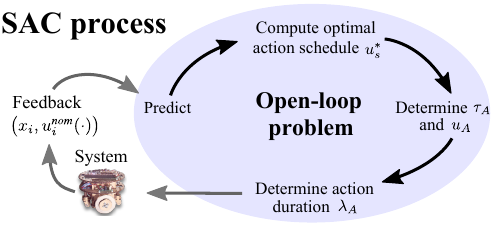}
\caption{An overview of the SAC control process. \deleted{One major advantage of SAC compared to other MPC approaches is that the open-loop problem in SAC---see \eqref{open_loop_problem}---can be solved without employing nonlinear programming solvers \cite{wachter2006implementation}.} In order to solve \eqref{open_loop_problem}, the SAC method follows the four steps shown above.}
  \label{fig:sac_process}
\end{figure}

For convenience, we briefly summarize the work presented in \cite{SACtro,SACexosekApplic,tzorakoleftherakis2016model,Fan-RSS-16}. SAC is a receding horizon method that enables real-time, closed-loop constrained control synthesis by following the cyclic diagram in Fig.~\ref{fig:sac_process}. We shall consider nonlinear systems with input constraints such that
\begin{gather}
\label{f}
\dot{x} = f(t,x,u) = g(t,x) + h(t,x) \, u \;\;\;\forall t \\
\text{with\;\;} u\in\mathcal{U} \;\;\text{and}\notag\\
\begin{aligned}
\mathcal{U}:=\Big\{ u\in\mathbb{R}^p: u_{\text{\emph{min}}}\leq u \leq u_{\text{\emph{max}}},\; u_{\text{\emph{min}}} \leq 0 \leq u_{\text{\emph{max}}}\Big\} \notag \text{,}
	\end{aligned}
\end{gather}
i.e., systems that can be nonlinear with respect to the state vector,  \mbox{$x:\mathbb{R}\to\mathcal{X}$}, but are assumed to be linear with respect to the control vector,  \mbox{$u:\mathbb{R}\to\mathcal{U}$}. \added{State constraints may be added in the form of penalty terms in the cost.} The state will sometimes be denoted as $t \mapsto x\big(t;x(t_i), u(\cdot)\big)$ when we want to make explicit the dependence on the initial state (and time), and corresponding control signal.

The SAC method uses objectives of the form 
%
\begin{equation}
\label{Jtrack}
J\big(x(\cdot)\big) = \int_{t_i}^{t_i+T} l\big(t,x(t)\big) \,dt + m\big(t_i+T,x(t_i+T)\big) \text{, }
\end{equation}
%
with incremental cost  \mbox{$l\big(t,x(t)\big)$}, terminal cost \mbox{$m\big(t_i+T,x(t_i+T)\big)$} and time horizon $T$. Although \eqref{Jtrack} lacks a norm on control effort, this norm is included in one of the subsequent steps as shown in \eqref{Ju}. The following definition is necessary before introducing the open-loop problem solved in SAC.
\begin{Definition}
\label{def1}
An action $A$ is defined by the triplet consisting of a constant control value, $u_A \in \mathcal{U}$, application duration, \mbox{$\lambda_A \in \mathbb{R^+}$} and application time, $\tau_A \in \mathbb{R}$, such that \mbox{$A:=\{u_A, \lambda_A, \tau_A\}$}. 
\end{Definition}

The open-loop problem in SAC calculates controls that \emph{improve} (not optimize) the objective \eqref{Jtrack} relative to applying only a nominal control signal. Specifically, the open-loop problem $\mathcal{P}$ in SAC is defined as follows 
\begin{align}
\label{open_loop_problem}
&\mathcal{P}(t_i,\,x_{i}):\\
 &\text{Find action } A \text{ such that} \notag \\
 &J\big(x(t;x_{i}, u_i^*(\cdot))\big)<J\big(x(t;x_{i}, u_i^{\text{\emph{nom}}}(\cdot))\big) \label{improve_condition}\\
&\text{subject to} \notag \\
&u_i^*(t) = \begin{cases} 
      u_{A} & \tau_A\leq t\leq \tau_A+\lambda_A \\
      u_i^{\text{\emph{nom}}}(t) & \text{else}
   \end{cases} \text{,} \notag\\
 & \tau_A >= t_i \text{,\,\,\;}\tau_A + \lambda_A <= t_i +T \text{,} \notag \\
&\text{and }\eqref{f} \text{ with } t\in[t_i,t_i+T] \text{ and }x(t_i)=x_{i} \notag
\phantom{\hspace{3cm}}
\end{align}
where $u_i^{\text{\emph{nom}}}$ is a nominal control signal \added{(see Remark~\ref{nom_def})}. \added{The subscript $i$ denotes the $i$-th time step, starting from $i=0$ and will be used for the remainder of the paper in the same way.}
\begin{figure}[t!]
\centering
\includegraphics[width=3.1in]{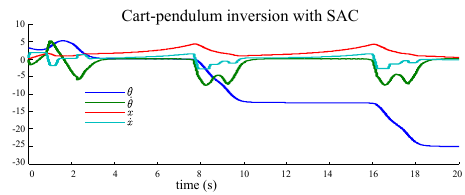}
\caption{\added{Cart-pendulum inversion with SAC. While SAC can drive the system close to the upward equilibrium, it does not achieve final stabilization.}}
  \label{fig:sac_cartpend}
\end{figure}

\begin{figure*}[t!]
  \centering
  \includegraphics[width=6.8in]{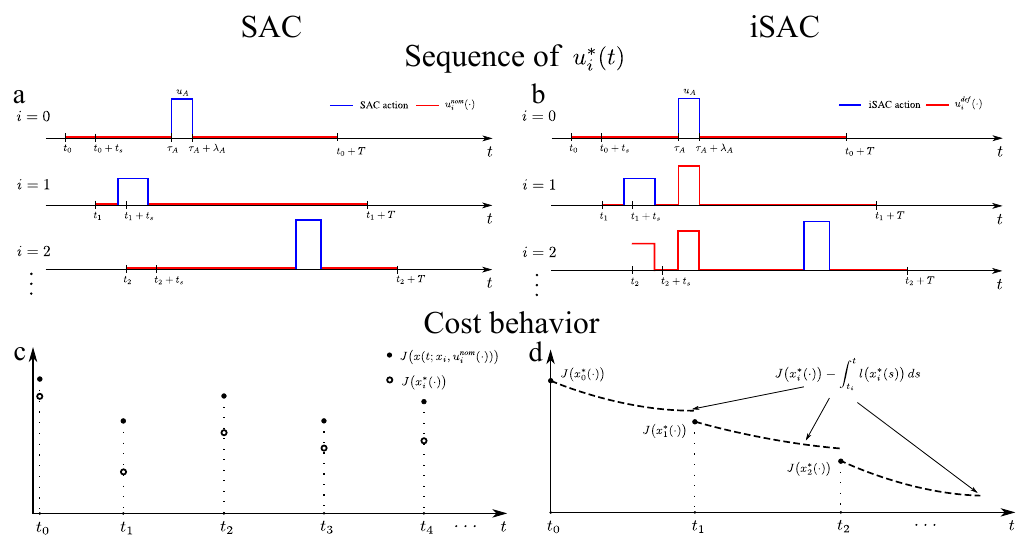}
  \caption{Summary of differences between SAC and iSAC. Panels a and b underline the fact that iSAC stores previously calculated actions and iteratively improves the open-loop solution $u_i^*(t)$ (assuming zero nominal control). As a result, a contractive constraint can be employed in iSAC to establish a sufficient decrease condition for the cost as shown in panel d. Panel c shows the corresponding graph in SAC as dictated by \eqref{improve_condition}.}
  \label{fig:sac_vs_isac}
\end{figure*}

The solution $u_i^*(t)$ of problem $\mathcal{P}$ includes a switch to the calculated action $A$ for \mbox{$t \in [\tau_A,\tau_A + \lambda_A]\subseteq[t_i,t_i+T]$} (Fig.~\ref{fig:sac_vs_isac}a) and thus, it is piecewise continuous in $t$. When applied to \eqref{f}, $u_i^*(t)$ generates a (discontinuous) switch of the same duration $\lambda_A$ in the dynamics from \mbox{$f\big(t,x,u_i^{\text{\emph{nom}}}(\cdot)\big)$} to \mbox{$f(t,x,u_A)$}. The condition in~\eqref{improve_condition} highlights a key feature of SAC, e.g., rather than optimize the objective~\eqref{Jtrack}, SAC actions \emph{improve} \eqref{Jtrack} relative to only applying nominal control \added{(see Remark~\ref{nom_def})}. As the receding horizon strategy progresses, \mbox{$\mathcal{P}(t_i, x_{i})$} is solved from the current time $t_i$ using the measured state $x_{i}$, and the calculated control $u_i^*(t)$---corresponding to $x_i^*(t)$---is applied for $t_s$ seconds (sampling time) to drive the system from $x_{i}$ to \mbox{$x\big(t_i+t_s;t_i, x_{i}, u_i^*(\cdot)\big)$}. The process is then repeated at the next sampling instance, i.e., \mbox{$t_i \leftarrow t_i+t_s$} and \mbox{$i \leftarrow i+1$}. The final result of the closed-loop receding horizon strategy is a sequence of actions, forming a piecewise continuous control signal $\overline{u}_{cl}(t)$ with state response $\overline{x}_{cl}(t)$. 

Unlike alternative MPC methods \cite{chen1998nonlinear,mayne2000constrained,fontes2001general,grune2011nonlinear,lee2011model,mayne2014model}, the open-loop problem in SAC can be solved in closed form without employing nonlinear programming solvers (see \cite{wachter2006implementation} and \cite{biegler2013survey} for a review). In order to solve \eqref{open_loop_problem}, the SAC method follows four steps as illustrated in Fig.~\ref{fig:sac_process}. Beginning with prediction, the steps of the SAC process are described in the following subsections.

\begin{Remark}
\label{nom_def}
\added{Nominal control $u^{\text{\emph{nom}}}: \mathbb{R} \rightarrow \mathcal{U}$ is piecewise continuous in $t$, and is used as a basis when calculating the open-loop solution. It is often $u^{\text{\emph{nom}}}(\cdot)\equiv 0$ so that problem $\mathcal{P} $ outputs the optimal action relative to doing nothing (allowing the system to drift). Alternatively, $u^{\text{\emph{nom}}}(\cdot)$ may be an optimized feedforward or state-feedback controller, e.g., regulating an unstable equilibrium. For purposes of evaluation, the nominal control throughout this paper is considered constant or zero. The system trajectory corresponding to application of nominal control will be denoted as $x\big(t;x(t_i), u^{\text{\emph{nom}}}(\cdot)\big)$ or $x^{\text{\emph{nom}}}(\cdot)$ for brevity.}
\end{Remark}
\subsection*{Steps for solving the open-loop problem $\mathcal{P}$}
The solution process assumes the following:
\begin{Assumption}
The control objective is to steer the state to the origin. This is not a restrictive assumption as most control scenarios (including trajectory tracking) can satisfy it with a change of coordinates.
\end{Assumption}

\begin{Assumption}
\label{f_assum}
The dynamics $f$ in \eqref{f} are continuous in $u$, piecewise continuous in $t$ and continuously differentiable in $x$.  Also, $f$ is compact, and thus bounded, on compact sets $\mathcal{X}$ and $\mathcal{U}$. Finally, the system is assumed to be controllable and \mbox{$f(\cdot,0,0)=0$}.
\end{Assumption}
\begin{Assumption}
\label{ml_assum}
The terminal cost $m$ is positive definite and continuously differentiable. The incremental cost $l(t,x)$ is continuous in $t$ and continuously differentiable in $x$ and \mbox{$l(\cdot,0)=0$}. Also, there exists a continuous positive definite, radially unbounded function \mbox{$M:\mathcal{X} \to \mathbb{R}^+$} such that \mbox{$l(t,x) \geq M(x) \; \forall t$}. 
\end{Assumption}
\begin{Assumption}
\label{bounded_traj}
The trajectory \mbox{$x_i^*(t) \in \mathcal{X}$} corresponding to the solution \mbox{$u_i^*(t) \in \mathcal{U}$} of $\mathcal{P}(t_i,x_i)$ is absolutely continuous, and thus bounded, in \mbox{$[t_i,t_i+T]$}. 
\end{Assumption}
The open-loop problem $\mathcal{P}$ can then be solved by following the four steps presented below. 
\subsubsection{Predict}
\label{prediction}
The SAC process begins by predicting the evolution of a system model from current state feedback. In this step, the system \eqref{f} is simulated from the current state $x_i$ and time $t_i$, under $u_i^{\text{\emph{nom}}}(t)$ for \mbox{$t \in [t_i,\;t_i+T]$}. The sensitivity of \eqref{Jtrack} to the state variations along the predicted trajectory $x_i^{\text{\emph{nom}}}(\cdot)$ is provided by an adjoint variable,  \mbox{$\rho_i:[t_i,\;t_i+T]\to\mathbb{R}^n$}, also simulated during the prediction step. The adjoint satisfies
\begin{gather}
\label{rhodot}
\dot \rho_i = -D_{2}l\big(t,x_i^{\text{\emph{nom}}}\big)^T - D_{2}f\big(t,x_i^{\text{\emph{nom}}},u_i^{\text{\emph{nom}}}\big)^T \rho_i \\
\text{subject to\;\;} \rho_i(t_i+T) = D_{2}m\big(t_i+T,x_i^{\text{\emph{nom}}}(t_i+T)\big)^T\notag \text{,}
\end{gather}
where $D_i$ denotes derivative with respect to $i$-th argument.
  
The prediction phase completes upon simulation of the state using \eqref{f} and the adjoint system \eqref{rhodot} under $u_i^{\text{\emph{nom}}}(\cdot)$ control. The resulting trajectories $x_i^{\text{\emph{nom}}}(\cdot)$, $\rho_i (\cdot)$ will be used in the remaining three steps of the solution process.

\subsubsection{Compute optimal action schedule $u^*_s(\cdot)$}
\label{computeoptimal}
\begin{figure}
\centering
\includegraphics[width=3.3in]{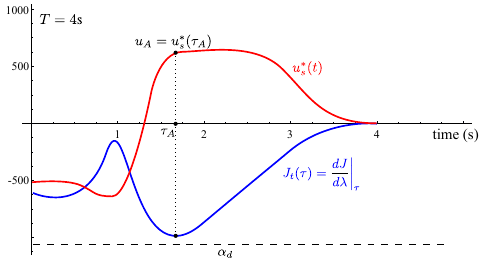}
\caption{A sample, one-dimensional ($m=1$) action schedule $u_s^*(t)$ and the corresponding cost \eqref{Jt} used to calculate the application time $\tau_A$ for the hypothetical time window \mbox{$[0, 4]$s}. Every point on $u_s^*(t)$ corresponds to an action of infinitesimal duration, with value and application time as determined by the curve. Because $u_s^*(t)$ is calculated from minimization of \eqref{Ju}, all actions in $u_s^*(t)$ would improve \eqref{Jtrack} relative to only applying nominal control if applied for $\lambda \to 0^+$. Each point on the mode insertion gradient curve approximates the change in cost \eqref{Jtrack} achievable by infinitesimal application of the corresponding action in $u_s^*(t)$. By choosing the application time that minimizes this curve, we pick the action the generates the greatest cost reduction in the current time window.}
  \label{fig:sac_sample}
\end{figure}

In this step, we compute a schedule \mbox{$u^*_s:[t_i, t_i+T] \to\mathbb{R}^p$} which contains candidate action values and their corresponding application times, assuming $\lambda \to 0^+$ for all (see Fig.~\ref{fig:sac_sample} for a sample one-dimensional action schedule). The final $u_A$ and $\tau_A$ will be selected from these candidates in step three of the solution process such that \mbox{$u_{A}=u_s^*(\tau_A)$}, while a finite duration $\lambda_A$ will be selected in step four. The optimal action schedule $u^*_s(\cdot)$ is calculated by minimizing 
\begin{gather}
\label{Ju}
J_{u_s} = \frac{1}{2} \int_{t_i}^{t_i+T} \bigg [\frac{dJ}{d \lambda}(t) - \alpha_d \bigg ]^2 + \lVert u_s(t)-u_i^{\text{\emph{nom}}}(t) \rVert_{R}^2 \,dt \text{,} \\
\frac{dJ}{d \lambda}(t)=\rho_i(t)^T \left[f\big(t,x_i^{\text{\emph{nom}}}(t),u_s(t)\big)-f\big(t,x_i^{\text{\emph{nom}}}(t),u_i^{\text{\emph{nom}}}(t)\big)\right] \notag 
\end{gather}
where the quantity $\frac{dJ}{d \lambda}(\cdot)$, called mode insertion gradient (see \cite{eger,egerstedt2006transition}), denotes the rate of change of the cost with respect to a switch of infinitesimal duration $\lambda$ in the dynamics of the system. In this case, $\frac{dJ}{d \lambda}(\cdot)$ shows how the cost will change if we introduce infinitesimal switches from \mbox{$f\big(t,x_i^{\text{\emph{nom}}}(t),u_i^{\text{\emph{nom}}}(t)\big)$} to \mbox{$f\big(t,x_i^{\text{\emph{nom}}}(t),u_s(t)\big)$} at any time \mbox{$t\in [t_i,t_i+T]$}. Intuitively, minimization of~\eqref{Ju} is driving $\frac{dJ}{d \lambda}(\cdot)$ to a negative value \mbox{$\alpha_d \in \mathbb{R}^-$}. As a result, each switch/action value in $u_s^*(t)$ is the single choice that improves \eqref{Jtrack} (relative to only applying nominal control) if applied for $\lambda \to 0^+$ at its corresponding application time. The design parameter $\alpha_d$ determines how much the cost is improved by each infinitesimal action in the schedule $u_s^*(t)$.

\begin{Remark}
\added{Equations \eqref{rhodot} and the mode insertion gradient in \eqref{Ju} can alternatively be explained as the adjoint equation and the variation of the Hamiltonian in Pontryagin's Maximum Principle (see, e.g., \cite{hale2016hamiltonian}). However, note that our method uses the variation of the Hamiltonian away from the optimizer, which can be verified by observing that \eqref{rhodot} depends on nominal state and control trajectories which are known a priori at each time step.}
\end{Remark}

Based on the simulation of the dynamics \eqref{f}, and \eqref{rhodot} completed in the prediction step (Section~\ref{prediction}), minimization of \eqref{Ju} leads to the following closed-form expression for the optimal action schedule:
%
%
\begin{equation}
\label{uopt}
u^*_s(t) = u_i^{\text{\emph{nom}}}(t)+(\Lambda + R^T)^{-1} \,  h\big(t,x_i^{\text{\emph{nom}}}(t)\big)^T \rho_i(t) \, \alpha_d \text{,}
\end{equation}
%
where \mbox{$\Lambda \triangleq h\big(t,x_i^{\text{\emph{nom}}}(t)\big)^T \rho_i(t) \rho_i(t)^T h\big(t,x_i^{\text{\emph{nom}}}(t)\big)$}. \added{Derivation of \eqref{uopt} is a straightforward extension of the proof of Theorem~1 in \cite{SACtro}. The only difference is that, here, we added the term $u_i^{\text{\emph{nom}}}(t)$ in the control norm in \eqref{Ju}, to get a more compact form for the solution \eqref{uopt}.}

The infinitesimal action schedule can then be directly saturated to satisfy any min/max control constraints of the form \mbox{$u_{\text{\emph{min,}}k} \leq 0 \leq u_{\text{\emph{max,}}k} \; \forall k \in \{1, \dots , m\}$} such that $u^*_s \in \mathcal{U}$ without additional computational overhead (see \cite{SACtro} for proof). This is possible because the SAC method does not optimize but rather improves \eqref{Jtrack}, and as such, even if the schedule is directly saturated, each action in the schedule will still point to a descent direction of the cost.

\added{The following assumption is used in conjunction with Proposition~\ref{negative_djdl} to ensure that \eqref{uopt} leads to negative mode insertion gradient, and thus, decreases the cost \eqref{Jtrack}.}
\begin{Assumption}
\label{actionability}
It is assumed that \mbox{$h\big(t, x(t)\big)^T\rho(t) \neq 0^{p \times 1}$} (related to controllability---see \cite{mamakoukasfeedback}). \added{Also, we assume that $u_i^{\text{\emph{nom}}}$ is not an optimizer of \eqref{Jtrack} in the current time step (usually selected as constant or zero---see Remark~\ref{nom_def}), and that $J\big(x_i^*(t)\big)\neq 0$, i.e., system trajectories have not already converged to the desired equilibrium.}
\end{Assumption}

\subsubsection{Determine application time $\tau_A$ (and thus $u_{A}$ value)}
\label{whentoact}
The SAC method optimizes a decision variable not normally included in control calculations---the choice of \emph{when} to act. As opposed to always acting at the current time, i.e., $\tau_A = t_i$, the application time of an action is allowed to take values in \mbox{$\tau_A \in [t_i,\;t_i+T]$}. 

Recall that the curve $u^*_s(\cdot)$ in the previous step provides the values and application times of possible infinitesimal actions that SAC could take at different times to improve \eqref{Jtrack} from that time. In this step the SAC method chooses one of these actions to apply, i.e., chooses the application time $\tau_A$ and thus an action value $u_{A}$ such that $u_{A}=u^*_s(\tau_A)$. To do that, $u^*_s(\cdot)$ is searched for a time $\tau_A$ that minimizes
\begin{gather}
\label{Jt}
J_{t}(\tau) = \frac{dJ}{d \lambda} \bigg \vert_{\tau} \text{,} \\
\frac{dJ}{d \lambda} \bigg \vert_{\tau}=\rho_i(\tau)^T \left[f\big(\tau,x_i^{\text{\emph{nom}}}(\tau),u^*_s(\tau)\big)-f\big(\tau,x_i^{\text{\emph{nom}}}(\tau),u_i^{\text{\emph{nom}}}(\tau)\big)\right] \notag \\
\text{subject to } \tau \in [t_i,t_i+T] \text{.} \notag
\end{gather}
Notice that the cost \eqref{Jt} is the mode insertion gradient evaluated at the optimal schedule $u_s^*(\cdot)$. Thus, minimization of \eqref{Jt} is equivalent to selecting the infinitesimal action from $u_s^*(\cdot)$ that will generate the greatest cost reduction relative to only applying nominal control. For a sample $J_{t}(\tau)$ see Fig.~\ref{fig:sac_sample}. \deleted{Finally, the selected application time must satisfy the following:}

\subsubsection{Determine control duration $\lambda_A$}
\label{howlong}
So far, $\tau_A$ and $u_A$ (see Definition~\ref{def1}) have been selected from a schedule of possible \emph{infinitesimal} actions, $u_s^*(\cdot)$. The final step in synthesizing a SAC action is to choose how long to act, i.e., a finite control duration $\lambda_A$, such that the condition in~\eqref{improve_condition} is satisfied. The following assumption and proposition will facilitate the analysis in the sequel.
\begin{Proposition}
\label{negative_djdl}
For a choice of \mbox{$\alpha_d<0$} in \eqref{Ju}, an infinitesimal control action $u_s^*(\tau)$ that is selected according to Assumption~\ref{actionability} will result in \mbox{$\frac{dJ}{d\lambda}(\cdot) < 0$}. 
\end{Proposition}
\begin{IEEEproof}
See Appendix.
\end{IEEEproof}

Proposition~\ref{negative_djdl} proves that if $\alpha_d<0$, then the infinitesimal $u_s^*(\tau_A)$ will lead to a negative $\frac{dJ}{d\lambda}(\cdot)$. From \cite{egerstedt2006transition,caldwell2012projection}, there is a non-zero neighborhood around $\lambda \to 0^+$ where the mode insertion gradient models the change in cost in \eqref{improve_condition} to first order, and thus, a finite duration $\lambda_A$ exists that satisfies \eqref{improve_condition}. In particular, for \emph{finite} durations $\lambda$ in this neighborhood we can write 
\begin{align}
\label{djdlam_first_order}
J\big(x(t;x_{i}, u_i^*(\cdot))\big)-J\big(x(t;x_{i}&, u_i^{\text{\emph{nom}}}(\cdot))\big) \notag\\
&=\Delta J \approx \frac{dJ}{d \lambda} \bigg \vert_{\tau_A} \lambda \text{.} 
\end{align}
Thus, from Proposition~\ref{negative_djdl} and \eqref{djdlam_first_order}, the condition in \eqref{improve_condition} is feasible. Then, a finite action duration $\lambda_A$ can be calculated by employing a \emph{line search} process \cite{caldwell2012projection}. Starting with an initial duration $\lambda_0$ we simulate the effect of the control action using \eqref{f} and \eqref{Jtrack}. If the simulated action satisfies~\eqref{improve_condition}, the duration is selected.  If this is not the case, the duration is reduced and the process is repeated. By continuity, the final duration will produce a change in cost within tolerance of that predicted from \eqref{djdlam_first_order}.


After computing the duration $\lambda_A$, the control action $A$ is fully specified (it has a value, an application time and a duration) and thus the solution $u_i^*(t)$ of problem $\mathcal{P}$ has been determined. By iterating on this process (Section~\ref{prediction} until Section~\ref{howlong}), SAC synthesizes piecewise continuous, constrained control laws for nonlinear systems. For more information about SAC, the reader is encouraged to consult \cite{SACtro,tzorakoleftherakis2016model,Fan-RSS-16}.

%

%% file: isac_desc.tex
\label{isac}

\added{In this section we will describe the differences between SAC and iSAC, and reformulate the open-loop problem that is being solved. We will focus on the two key modifications introduced in iSAC, i.e., a) the iterative update of the open-loop solution $u_i^*(t)$ across time steps, and b) the application of a contractive constraint on the cost. These two features are illustrated in Fig.~\ref{fig:sac_vs_isac} and are discussed in more detail in the following paragraphs.} For convenience, the corresponding SAC behavior is also shown in the same figure.
\subsection{Iterative update of $u_i^*(t)$}
Figure~\ref{fig:sac_vs_isac}a illustrates how the iSAC method stores actions calculated at previous time steps and modifies $u_i^*(t)$ by introducing a new action at every time step. On the contrary, SAC only keeps a single action in $u_i^*(t)$, regardless of whether that action lies in the current application window \mbox{$[t_i, t_i+t_s]$} or not. In light of this, the following definition is necessary to distinguish between the nominal control (zero or constant in this paper) from the open-loop problem $\mathcal{P}$ in SAC, and the default control that iSAC uses to calculate actions.

\begin{Definition}
Default control \mbox{$u_i^{\text{\emph{def}}}: [t_i, t_i+T] \rightarrow \mathcal{U}$}, is piecewise continuous in $t$ and is defined as 
\begin{align}
\label{default_control}
u_i^{\text{\emph{def}}}(t) = \begin{cases} 
      u_{i-1}^*(t) & t_i \leq t\leq t_i+T-t_s \\
      u_i^{\text{\emph{nom}}}(t) & t_i+T-t_s < t\leq t_i+T
   \end{cases} \text{,}
\end{align}
\end{Definition}
with \mbox{$u_0^{def}(\cdot) \equiv u_0^{\text{\emph{nom}}}(\cdot)$}. In expression \eqref{default_control}, $u_{i-1}^*: [t_{i-1},t_{i-1}+T] \rightarrow \mathcal{U}$ is the output  of  $\mathcal{P}(t_{i-1}, x_{i-1})$ from the previous time step $i-1$, and    $t_s=t_i-t_{i-1}$ is the sampling period. 

As a result, actions calculated in iSAC improve the cost~\eqref{Jtrack} with respect to applying default control $u_i^{\text{\emph{def}}}(\cdot)$. Specifically, as shown in Fig.~\ref{fig:sac_vs_isac}d, iSAC actions establish a decreasing behavior\footnote{In Section~\ref{stability} we make use of this fact and consider~\eqref{Jtrack} as a candidate Lyapunov function.} for \eqref{Jtrack} from time step to time step.  As will be explained in the following paragraph, this is accomplished by applying a contractive constraint on the cost, while the iterative nature of $u_i^*(t)$ has a major role in the formulation of this constraint.

\subsection{Contractive constraint on cost}

In SAC, each action is calculated such that \eqref{improve_condition} is satisfied. However, in general, this does not ensure that the cost will follow a decreasing trend. The explanation lies in the fact that \eqref{improve_condition} only guarantees that the cost will improve relative to a nominal input in the current time step, but not necessarily across time steps. As an example, Fig.~\ref{fig:sac_vs_isac}c shows a general scenario where condition \eqref{improve_condition} is met. In this case, the system does not meet the desired specifications encoded in the cost, since the latter is not decreasing with time. Thus, even if \eqref{improve_condition} is satisfied, closed-loop stability cannot be established in SAC.

A solution to this problem is to modify condition \eqref{improve_condition} such that \eqref{Jtrack} sufficiently ``contracts" between time steps \cite{de2000contractive,camponogara2002distributed,xie2008first,ferrari2009model}. \added{Contractive constraints have been widely used in the MPC literature to show closed-loop stability as an alternative to methods relying on a terminal (region) constraint \cite{chen1998nonlinear,mayne2000constrained,fontes2001general,grune2011nonlinear,lee2011model,mayne2014model}. A disadvantage of the latter is that they require the computation of a terminal region, which has to be calculated separately for each system of interest \cite{chen2003terminal,yu2009enlarging,SOSderivParrilo2000}. On the contrary, the contractive constraint approach can be applied to a variety of examples without modification and is easier to implement.}

 Specifically, instead of improving the cost relative to only applying nominal control in \eqref{improve_condition}, we apply 
\begin{equation}
\label{new_improve}
J\big(x_i^*(\cdot)\big) - J\big(x_{i-1}^*(\cdot)\big) \leq-\int_{t_{i-1}}^{t_{i}} l\big(t, x_{i-1}^*(t)\big) \,dt\text{, } 
\end{equation} 
as a contractive constraint. Conditions similar to \eqref{new_improve} also appear in terminal region methods, either in continuous or in discrete time, as an intermediate step used to prove closed-loop stability.

One problem that arises with \eqref{new_improve} in SAC is that the quantities \mbox{$J\big(x_i^*(\cdot)\big)$} and \mbox{$J\big(x_{i-1}^*(\cdot)\big)$} cannot be related through the mode insertion gradient, unlike, e.g., \mbox{$J\big(x_{i}^*(\cdot)\big)$} and \mbox{$J\big(x_{i}^{\text{\emph{nom}}}(\cdot)\big)$} that appear in the original condition \eqref{improve_condition} and can be related through \eqref{djdlam_first_order}. Thus, Proposition~\ref{negative_djdl} can no longer be used to ensure that \eqref{new_improve} is feasible in SAC.
%
%
On the contrary, in iSAC, we can use the iterative nature of $u_i^*(t)$ to ensure feasibility of \eqref{new_improve} through Proposition~\ref{negative_djdl} for sufficiently small sampling time $t_s$. In particular, in iSAC we can  transform \eqref{new_improve} to an equation that is similar to \eqref{improve_condition} and includes the terms $J\big(x_{i}^*(\cdot)\big)$ and $J\big(x_{i}^{\text{\emph{def}}}(\cdot)\big)$, which can be related through the mode insertion gradient $\frac{dJ}{d \lambda}$, as in \eqref{djdlam_first_order}. 

By definition, \eqref{default_control} leads to $x_{i}^{\text{\emph{def}}}(t) \equiv x_{i-1}^*(t)$ for $t \in [t_i, t_{i-1}+T]$ (see also Fig.~\ref{fig:sac_vs_isac}b). We can then write
\begin{align}
&J\big(x_{i-1}^*(\cdot)\big)-J\big(x_{i}^{\text{\emph{def}}}(\cdot)\big) = \int_{t_{i-1}}^{t_{i}} l\big(t, x_{i-1}^*(\cdot)\big) \,dt  \notag \\
&+ m\big(t_{i-1}+T,x_{i-1}^*(t_{i-1}+T)\big) 
- \int_{t_{i-1}+T}^{t_{i}+T} l\big(t, x_{i}^{\text{\emph{def}}}(t)\big) \,dt \notag\\ &-m\big(t_{i}+T,x_{i}^{\text{\emph{def}}}(t_{i}+T)\big)\text{.}
 \label{suff_decrease1}
\end{align}
Combining \eqref{suff_decrease1} with \eqref{new_improve} and using \eqref{default_control} we get
%
%
\begin{align}
\label{final_isac_improv_init}
& J\big(x_i^*(\cdot)\big) -  J\big(x_{i}^{\text{\emph{def}}}(\cdot)\big) \leq m\big(t_{i-1}+T,x_{i-1}^*(t_{i-1}+T)\big) \notag \\
&- \int_{t_{i-1}+T}^{t_{i}+T} l\big(t, x_{i}^{\text{\emph{def}}}(t)\big) \,dt
-m\big(t_{i}+T,x_{i}^{\text{\emph{def}}}(t_{i}+T)\big)
\end{align}
or equivalently
\begin{align}
\label{final_isac_improv}
J\big( & x_i^*(\cdot)\big) - J\big(x_{i}^{\text{\emph{def}}}(\cdot)\big) \leq \notag \\
&- \int_{t_{i-1}+T}^{t_{i-1}+t_s+T} l\big(t, x_{i}^{\text{\emph{def}}}(t)\big) + \dot{m}\big(t,x_{i}^{\text{\emph{def}}}(t)\big) \,dt = C \text{.}
\end{align}
Thus, we were able to transform \eqref{new_improve} into a sufficient decrease condition similar to \eqref{improve_condition}, i.e., a condition that involves $J\big(x_{i}^*(\cdot)\big)$ and $J\big(x_{i}^{\text{\emph{def}}}(\cdot)\big)$. \added{In terminal region methods, the integrand quantity in \eqref{final_isac_improv} is often required to be negative in some region of the state space and for some nominal control signal (see e.g. \cite{fontes2001general,mayne2000constrained}) to achieve closed-loop stability. Applying such a requirement in our case would make satisfaction of \eqref{final_isac_improv} trivial as the right-hand side of \eqref{final_isac_improv} would be positive and the left-hand side would be negative (Proposition~\ref{negative_djdl}). However, calculation of the terminal region and control is not straightforward and is also not necessary for iSAC.} The following Proposition can be used in conjunction with Proposition~\ref{negative_djdl} to ensure that the new condition \eqref{final_isac_improv} is (recursively) feasible. 
\begin{Proposition}
\label{exists pair}
%
For $\alpha_d<0$ there exists a sufficiently small sampling time $t_s$ such that the sufficient cost decrease required by the contractive constraint \eqref{final_isac_improv} is attainable. 
\end{Proposition}
\begin{IEEEproof}
See Appendix.
\end{IEEEproof}
The contractive constraint \eqref{final_isac_improv} can be applied in the line search process that determines the duration of an action (Section~\ref{howlong}) in lieu of \eqref{improve_condition}.
By applying this condition at each time step, we can guarantee that the cost value will decrease across time steps, and using this fact we can prove closed-loop stability (Section~\ref{stability}).

\subsection*{Open-loop problem in iSAC}
Based on the above, the open-loop problem solved by iSAC at each time step is given by
\begin{align}
\label{open_loop_problem_isac}
&\mathcal{B}(t_i,\,x_{i}):\\
 &\text{Find action } A \text{ such that } 
 \text{\eqref{final_isac_improv} is satisfied} \notag\\
&\text{subject to} \notag \\
&u_i^*(t) = \begin{cases} 
      u_{A} & \tau_A\leq t\leq \tau_A+\lambda_A \\
      u_i^{\text{\emph{def}}}(t) & \text{else}
   \end{cases} \text{,} \notag\\
    & \tau_A >= t_i \text{,\,\,\;}\tau_A + \lambda_A <= t_i +T \text{,} \notag \\
&\text{and }\eqref{f} \text{ with } t\in[t_i,t_i+T] \text{ and }x(t_i)=x_{i} \text{.} \notag
\phantom{\hspace{3cm}}
\end{align}
Similar to the open-loop problem $\mathcal{P}$ in SAC, the solution $u_i^*(t)$ of $\mathcal{B}$ can be obtained in closed form by following the four steps in Fig.~\ref{fig:sac_process} and Section~\ref{prelims}, without relying on nonlinear programming solvers. The only difference is that in the solution process of $\mathcal{B}$, the superscripts \emph{nom} are replaced by \emph{def}. Our final result in this section utilizes the aforementioned assumptions and propositions to ensure that the open-loop problem $\mathcal{B}$ has a solution:
\begin{Proposition}[Existence of solution to $\mathcal{B}$]
\label{existence_of_solution}
For sufficiently small sampling time $t_s$, the solution $u_i^*(t)$ to the open-loop problem $\mathcal{B}(t_i,\,x_{i})$ exists for any $x_i$, $t_i$.
\end{Proposition}
\begin{IEEEproof}
See Appendix.
\end{IEEEproof}
%

%% file: stability.tex
\label{stability}
In this section, we provide global stability results for iSAC and discuss how iSAC can be used under a special case of disturbances and to track infeasible trajectories, i.e., trajectories that do not satisfy the dynamics \eqref{f}.

\subsection{Nominal Case}
In this section we are considering nominal stability (not robust stability), i.e., the trajectories of the plant are assumed to coincide with the trajectories predicted by the model.  Recall that the iSAC method does not optimize the cost \eqref{Jtrack} but rather improves it across time steps. It is often the case that minimization of the objective function in the open-loop optimization problem is not required to achieve stability of a model predictive controller. Sometimes, a decrease in the cost at every iteration is sufficient to guarantee stability \cite{chen1997quasi,fontes2001general,scokaert1999suboptimal,jadbabaie2001unconstrained}.  In view of these observations, we now present the following Theorem:
\begin{Theorem}[Asymptotic Stability]\label{theorem1}For sufficiently small sampling time $t_s$, the closed-loop system resulting from applying iSAC is asymptotically stable in the sense that \mbox{$||\overline{x}_{cl}(t)||\to 0$ as $t\to \infty$. }
\end{Theorem}
\begin{IEEEproof}
See Appendix.
\end{IEEEproof}
Theorem~\ref{theorem1} does not imply Lyapunov stability, but rather establishes the usual notion of attractiveness. The importance of $t_s$ in establishing closed-loop stability is also underlined; if the assumptions of the Theorem hold, by appropriate selection of $t_s$, we can ensure that the contractive constraint is feasible, and thus, asymptotic stability.


\subsection*{Cost as a Lyapunov function}
Now that asymptotic stability has been proved, we will show that, under certain assumptions, the objective \eqref{Jtrack} is a Lyapunov function for the closed-loop system that decreases at intervals of prediction horizons. 

\begin{Theorem}[Objective Function as a discrete-time Lyapunov Function]\label{cost_lyap}Assume that $f$ is bounded such that \mbox{$||f|| \leq \xi ||x||$} for some finite constant $\xi > 0$, and that the cost \eqref{Jtrack} is of the form
\begin{equation}
\label{quadr_J_lyap}
J\big(t_i, x(\cdot)\big) = \frac{1}{2} \int_{t_i}^{t_i+T} ||x(t)||^2_Q \,dt + \frac{1}{2} ||x(t_i+T)||_{P_1}^2 \text{,}
\end{equation}
with $t_i$ taking values in a discrete set for $i \in \mathbb{N}$. Then, for sufficiently small $t_s$, \eqref{quadr_J_lyap} is a discrete-time Lyapunov function for the closed-loop system.
\end{Theorem}
\begin{IEEEproof}
The proof can be found in the Appendix.
\end{IEEEproof}
\begin{Remark}
\added{Theorem~\ref{cost_lyap} proves that a quadratic cost may be used as a discrete-time Lyapunov function in our method. The Lyapunov function does not have to be continuously decreasing along solution trajectories, thus only establishing attractiveness similar to Theorem~\ref{theorem1}. Additionally, even though \eqref{quadr_J_lyap} is continuously differentiable, there is no assumption on the continuity with respect to the state. These points are important, since they allow application of this theorem to problems that potentially do not admit a continuous-time Lyapunov function which is also continuous in the state, e.g. nonholonomic systems \cite{acc2003discontinuous}.}
\end{Remark}

\subsection{Asymptotically Decaying Disturbances}
This section provides stability results for a special case of disturbances. Here, we assume that the plant is described by the following differential equations (instead of \eqref{f})
\begin{gather}
\label{fp}
\dot{x}^p = f(t,x^p,u) + \eta(t) = g(t,x^p) + h(t,x^p) \, u +\eta(t)\text{,}
\end{gather}
where $x^p(t)$ represents the trajectory resulting from applying control $u$ and the effect of an unknown additive disturbance $\eta(t)$. The disturbance is bounded and asymptotically decaying; in particular, we assume the following:
\begin{Assumption}
\label{bounded_eta}
The disturbance $\eta(t)$ is bounded, i.e., $||\eta_i(t)||\leq \delta_i < \infty$ for all $t \in [t_i,t_i+T]$ and all $i>0$. Furthermore, $\eta(t)$ is asymptotically decaying, i.e., $\delta_i \to 0$ as $i \to \infty$.
\end{Assumption}
Closed-loop stability under $\eta(t)$ is provided by the following theorem.
\begin{Theorem}[Asymptotic Stability under Asymptotically Decaying Disturbances]
\label{theorem_dist}For sufficiently small $t_s$, the closed-loop system under the disturbance $\eta(t)$ is asymptotically stable in the sense that \mbox{$||\overline{x}_{cl}(t)||\to 0$ as $t\to \infty$. }
\end{Theorem}
\begin{IEEEproof}
The proof can be found in the Appendix.
\end{IEEEproof}
As seen in the proof, for as long as the disturbance is acting, the cost is allowed to increase. As time progresses, the disturbance attenuates and the contractive constraint becomes feasible again.

\subsection{Remarks on tracking of non-admissible trajectories}
\label{infeasible}
When designing a control task, there are often cases where asymptotic stability/convergence to the desired equilibrium cannot be achieved. For example, in trajectory tracking, it is not always feasible to identify trajectories that satisfy the dynamics of the system of interest, unless there is special structure one can exploit, e.g., differential flatness \cite{mellinger2011minimum}. The control goal in this case is to ensure that the deviation of the generated (feasible) open-loop trajectory from the desired one is bounded, i.e., $||x_i^*(t)-x^d(t)||\leq\Delta_i$ for $t \in [t_i, t_i+T]$ with $\Delta_i \in [0,\infty)$. Unlike the decaying disturbance case, the upper bound $\Delta_i$ does not have any structure that can be exploited. Thus there is no guarantee that $\overline{x}_{cl}(t) \to x^d(t)$ as $t\to\infty$. Nevertheless, even for this scenario, our simulation results show that iSAC keeps the deviation from the desired infeasible trajectories bounded.

\deleted{Similar to the disturbance case, these observations stem from the fact that the feasibility of the contractive constraint, and thus of the open-loop problem, can no longer be ensured when tracking an infeasible trajectory. In these cases, in order to keep the error from the desired behavior bounded, one can set $\alpha_d$ to more negative values such that the control generation is more ``aggressive". By doing so, the selected action leads to smaller cost increase (at a higher control cost) and iSAC generates a feasible response that is close to the desired infeasible objective, as shown in the simulations that follow.}

%% file: sim_results.tex
\label{simulationresults}
In this section we demonstrate the flexibility and versatility of iSAC through application to a variety of challenging systems. \added{For each example, we include literature review on related control methods, and we also compare the performance of iSAC on NASA's TRACE spacecraft \cite{karpenko2011flight} to pseudospectral infinite-horizon control \cite{fahroo2008pseudospectral,bedrossian2009zero}.} It is important to note that unlike other system-specific methods mentioned in this section, our approach requires no modifications across examples; only the dynamic model that is being controlled and the relevant parameters of iSAC are changed for each example. Also, all the examples presented here run much faster than real time, indicating applicability to hardware implementations. For better visualization of the results, the reader is encouraged to visit \url{https://vimeo.com/219702474}.

\begin{figure*}[t!]
  \centering
  \includegraphics[width=6.8in]{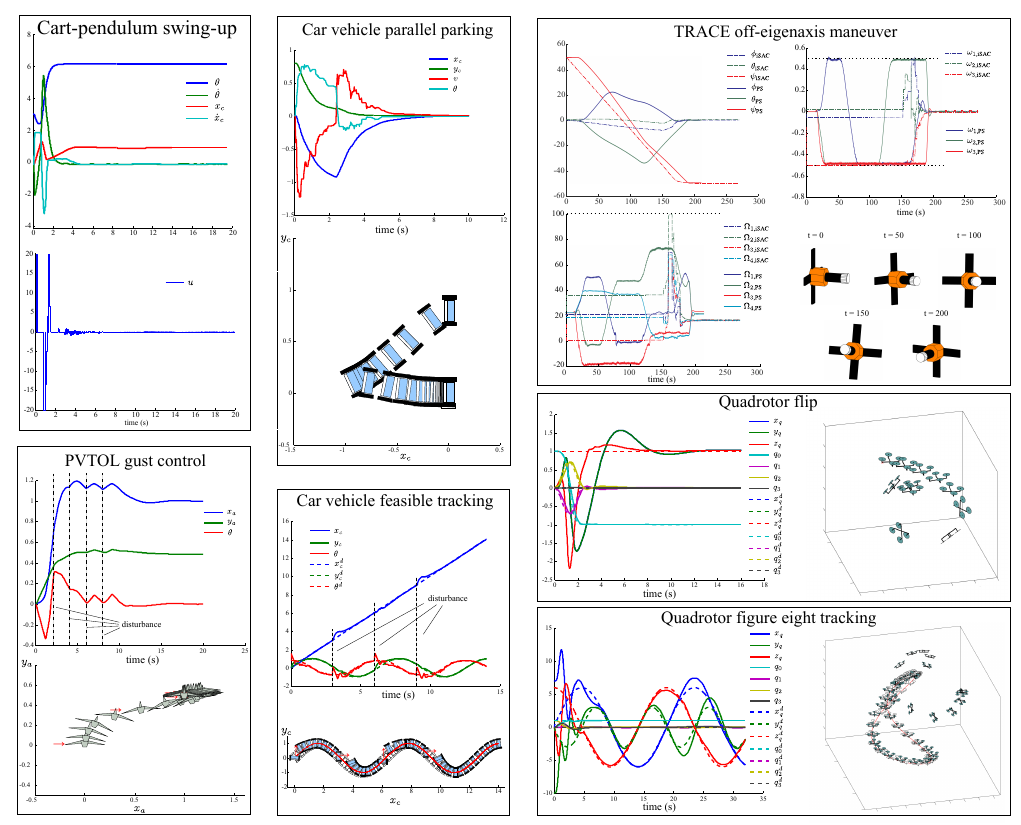}
  \caption{Summary of results; the iSAC method stabilizes the presented systems successfully. \deleted{An example of a closed-loop control generated by iSAC is provided in the acrobot panel, that also illustrates the linear state-feedback version of the method as in \eqref{u_linear}. The rest of the examples use only iSAC, without switching to \eqref{u_linear}.}Notice that our method is effective even in the presence of decaying disturbances and, also, keeps the deviation from the desired trajectory bounded in infeasible tracking. For better visualization of the results, the reader is encouraged to visit https://vimeo.com/219702474.}
  \label{fig:results}
\end{figure*}

The control objective is encoded using a quadratic cost throughout this section (Assumption~\ref{ml_assum} is thus satisfied). \added{In the TRACE example, we also incorporate state constraints by introducing penalty terms in the cost.} The examples also include cases of feasible and infeasible tracking, as well as disturbance rejection. The sampling time $t_s$ for each example was appropriately selected to achieve closed-loop stability (Theorems~\ref{theorem1} and~\ref{theorem_dist}) when feasible. Finally, a summary of iSAC-specific parameters for each example is given in Table~\ref{sac_params}, while Fig.~\ref{fig:results} shows the resulting trajectories for the examples described below. Our method is shown to be successful in these scenarios, leading to asymptotic stabilization when feasible.

\deleted{The dynamics of the system and the parameters used in the simulations that follow can be found in \cite{tzorakoleftherakis2015controllers}. The state vector for the system includes the two joint angles and their corresponding velocities, $[\theta_1, \dot{\theta}_1, \theta_2, \dot{\theta}_2]$. The control input is the torque at the distal joint.}

\deleted{The swing-up problem for the acrobot is a benchmark problem in control theory, where the system starts from the downward equilibrium and the control objective is stabilization at the upward equilibrium \cite{xin2012energy,spong1995swing}. The desired setpoint was thus $x^d=[0,0,0,0]$. To illustrate the duration-independent version of iSAC presented in Section~\ref{local_stability}, control calculation was based on Theorem~\ref{sac_lqr} after system trajectories were within a pre-defined region of the upward equilibrium, found also in \cite{tzorakoleftherakis2015controllers}. The cost weights were selected as \mbox{$Q = Diag[1000\,, 0\,, 500\,, 11]$}, and \mbox{$R = 0.01$}.}

\deleted{Figures~\ref{fig:results} and~\ref{fig:costs} show the resulting trajectories, closed-loop control and cost generated by iSAC. In contrast with other methods that switch between separately derived controllers \cite{albahkali2009swing}, iSAC pumps energy into the system by appropriate excitation, without explicitly encoding this behavior in the objective. Finally, many methods (see \cite{xin2012energy,xin2007analysis}) rely on energy-based objectives to solve the swing-up problem for the acrobot and pendulum systems, primarily because state-tracking objectives result in many local minima that prevent convergence to desirable trajectories. It is noteworthy that, like SAC, iSAC is able to invert the system using a state-tracking objective while bypassing local minima. For a  detailed comparison between SAC, sequential quadratic programming (SQP) \cite{gill1981practical}, and iLQG \cite{tassa2012synthesis} the reader is encouraged to consult \cite{SACtro} and \cite{Fan-RSS-16}.  }

\subsection{Cart-pendulum inversion}
\label{cartstep}
The cart-pendulum inversion is a benchmark problem in control theory \cite{tao2008design,aastrom2000swinging}. The dynamics of the system and the parameters used in the simulations that follow can be found in \cite{tzorakoleftherakis2015controllers}. The state vector for the system includes the angle and angular velocity of the pendulum and the horizontal position and velocity of the cart, $[\theta, \dot{\theta}, x_c, \dot{x}_c]$. The control input is the horizontal acceleration of the cart \added{with zero $u^{\text{\emph{nom}}}$}. The desired setpoint in this example was $x^d=[0,0,1,0]$. The cost weights were selected as \mbox{$Q = Diag[20\,, 0\,, 5\,, 0]$, $P = Diag[0.1\,, 0\,, 5\,, 0]$} and \mbox{$R = 0.3$}.

\added{Comparing the cart-pendulum response in Fig.~\ref{fig:results} with Fig.~\ref{fig:sac_cartpend}, it is clear that iSAC, unlike SAC, can achieve final stabilization. The proposed method is able to solve the inversion problem by pumping energy into the system without explicitly encoding this behavior in the objective. In contrast, alternative methods switch between separately derived controllers \cite{albahkali2009swing} to achieve the same result. Also, many methods (see \cite{xin2012energy,xin2007analysis}) rely on energy-based objectives to solve the swing-up problem for the acrobot and pendulum systems, primarily because state-tracking objectives result in many local minima that prevent convergence to desirable trajectories. It is noteworthy that, iSAC is able to invert the system using a state-tracking objective while bypassing local minima. The corresponding cost for this example is shown in Fig.~\ref{fig:costs}.}

\deleted{The desired trajectory in this example was selected as \mbox{$\theta^d(t)=\frac{\pi}{5} \sin t$} and the cost weights were \mbox{$Q = Diag[5000\,, 0\,, 0\,, 0]$, $P = Diag[500\,, 0\,, 0\,, 0]$}, \mbox{$R = 0.3$}. The desired trajectory was feasible given that no constraints were applied to the cart position and velocity. As the plot in Fig.~\ref{fig:results} illustrates, tracking was very accurate, making it hard to distinguish between the desired and actual trajectory followed by the system. The corresponding cost is shown in Fig.~\ref{fig:costs}.
As mentioned in the acrobot example, it is worth noting that iSAC is able to solve the inversion problem by pumping energy into the system without relying on separate controllers. In addition, this was accomplished using the state tracking objective, as opposed to the energy objective commonly employed (see for example \cite{aastrom2000swinging}), bypassing local minima that cause solutions to converge to undesired trajectories in state tracking costs.}

\subsection{Car-like vehicle}
The dynamics used in the simulations are
\begin{align}
\label{dyn}
&\dot{x}_c = v \cos \theta \text{,  } \dot{y}_c = v \sin \theta \text{,  }  \dot{v} = u_1  \text{,  }\dot{\theta} = u_2  \text{,}
\end{align}
where $x_c$, $y_c$ denote the position of the car, $\theta$ is the angle with respect to the horizontal axis and $v$ is the forward velocity. The iSAC method directly controls the acceleration and the angular velocity of the car. This nonholonomic system violates Brockett's necessary condition for smooth or even continuous stabilization \cite{brockett1983asymptotic}, which makes the control design problem challenging. Since iSAC automatically generates a discontinuous control law, it is not subject to this condition. 
\subsubsection{Parallel parking}\hfill
\label{carstep}

The desired setpoint for the parallel parking problem was \mbox{$x^d=[0,0,0,0]$} and the cost weights were selected as \mbox{$Q = Diag[1\,, 15\,, 0.8\,, 0.8]$, $P = Diag[0\,, 25\,, 0\,, 0]$} and \mbox{$R = Diag[0.1\,, 0.1]$}. As seen in Fig.~\ref{fig:results}, iSAC successfully drives the system to the origin. \added{Nominal control $u^{\text{\emph{nom}}}$ was set to zero for both inputs.} The corresponding cost is in Fig.~\ref{fig:costs}.

\subsubsection{Feasible trajectory tracking under disturbance}\hfill
\label{cartrack}

\begin{figure*}
  \centering
  \includegraphics[width=6.8in]{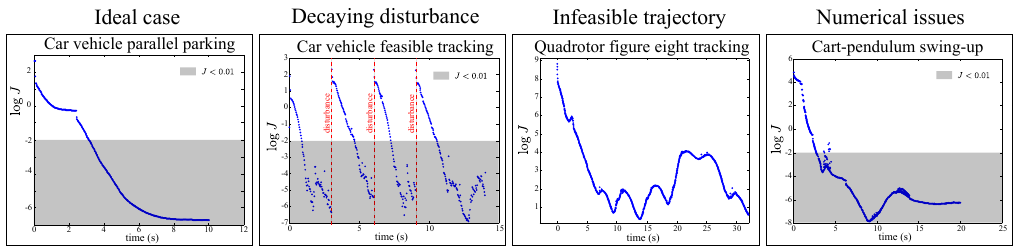}
  \caption{\replaced{Sample cost behavior in the presented examples. As explained in Section~\ref{sec:limitations}, due to numerical issues, the contractive constraint may not be satisfied momentarily near the desired equilibrium, leading to a small cost increase. Nevertheless, the system is effectively already stabilized at that point (compare corresponding trajectories from Fig.~\ref{fig:results}). Columns two and three show the cost behavior when acting under the presence of an external bounded disturbance and tracking infeasible trajectories respectively, where the cost is not expected to continuously decrease.}{Cost behavior in the presented examples. As explained in Section~\ref{sec:limitations}, numerical issues may arise when the system trajectories are near the desired equilibrium (see e.g., gray area in the plots). As a result, the contractive constraint may not be satisfied momentarily leading to a small cost increase. While this may not be a problem since the system is effectively already stabilized at that point (compare corresponding trajectories from Fig.~\ref{fig:results}), one way to deal with this issue is to switch to the duration-independent, linear-in-state version of iSAC in \eqref{u_linear} near the equilibrium (Theorem~\ref{sac_lqr}) as demonstrated by the acrobot swing-up panel. Finally, columns three and four show the cost behavior when acting under the presence of an external bounded disturbance and tracking infeasible trajectories respectively. For the former, the unmodeled disturbance leads to a temporary cost increase, but our method responds accordingly as the disturbance attenuates. For the latter, as explained in Section~\ref{infeasible}, the contractive constraint is no longer feasible which explains why the cost is not consistently decreasing. Nevertheless, iSAC keeps the deviation from the desired trajectory, e.g., the cost, bounded.}}
  \label{fig:costs}
\end{figure*}

The desired trajectory in this example was chosen as \mbox{$\big(x_c^d(t), y_c^d(t), \theta^d(t)\big)=(t, \sin t, \tan^{-1} \cos t)$} and the cost weights were \mbox{$Q = Diag[100\,, 100\,, 0\,, 10]$} and \mbox{$R = Diag[0.1\,, 0.1]$}. \added{Nominal control $u^{\text{\emph{nom}}}$ was set to zero for both inputs.} External disturbances acted instantaneously on the system on four occasions, i.e., at $t=0, 3, 6,$ and $9$s, each perturbing $v$ and $\theta$ by $+3\,\text{m}/\text{s}$ and $+\frac{\pi}{4}\,\text{rad}$ respectively. The effect of the disturbance can be seen in the corresponding plot in Fig.~\ref{fig:results}; while the car was tracking the desired trajectory, the disturbance pushed the system away every time it acted on the system. After the last disturbance attenuated the control successfully steered the system back to the desired trajectory. The corresponding cost is in Fig.~\ref{fig:costs}.

A great deal of work on nonholonomically constrained car-like models was completed in the 1990's using nonsmooth or time-varying control laws aiming to overcome stabilizability limitations traditional techniques  \cite{samson1991feedback,de1992exponential,astolfi1994stabilization}. Alternative methods that achieve similar results in the parking problem include fuzzy controllers as in \cite{chiu2005fuzzy} and dynamic feedback linearization \cite{de2000stabilization}. Other approaches that intrinsically lead to piecewise continuous controls like iSAC include MPC as in \cite{de2000contractive,gu2005stabilizing}. The former paper utilizes contractive constraints on the state and, although successful, results in a straight line path between the initial and final configuration. The latter uses the aforementioned terminal region constraints to achieve stabilization, but, unlike iSAC, the method is designed specifically for the car-vehicle system.

\subsection{Gust control of planar vertical take-off landing (PVTOL) aircraft}
\label{pvtol_gust}
The dynamic model used in this example is given in \cite{hauser1992nonlinear}. The state vector for the system includes the position of the aircraft, the angle with respect to the horizontal axis and the corresponding velocities, i.e., $[x_a, \dot{x}_a, y_a, \dot{y}_a, \theta, \dot{\theta}]$. The control inputs of the system are thrust (directed out the bottom of the aircraft) and the rolling moment. Lastly, the coupling parameter $\epsilon$ that appears in the model was chosen as $\epsilon=0.3$. Note that $\epsilon$ couples the rolling moment input with the lateral acceleration of the aircraft; a positive value $\epsilon>0$ means that applying a (positive) moment to roll left produces an acceleration to the right, making this a non-minimum phase system.

\added{Nominal control $u^{\text{\emph{nom}}}$ was set to 1N for the thrust and zero for the rolling moment.} The desired setpoint in this example was \mbox{$x^d=[1,0,0.5,0,0,0]$} and the cost weights were selected as \mbox{$Q = Diag[15\,, 3\,, 15\,, 3\,, 3\,, 0]$} and \mbox{$R = Diag[0.1\,, 0.1]$}. External disturbances acted instantaneously on the system on five occasions, i.e., at $t=0, 2, 4, 6,$ and $8$s, each perturbing $\dot{x}_a$ by $+0.1\,\text{m}/\text{s}$. The behavior of the system was similar to the one observed in the car-like vehicle; while the control was steering the system towards the desired setpoint, the disturbance drove it away until the last disturbance. 

Figure~\ref{fig:results} shows the resulting trajectories. The control of PVTOL aircrafts has been extensively studied using, e.g., gain-scheduling \cite{wu2008gain}, robust control \cite{lin1999robust} and input-output linearization \cite{hauser1992nonlinear}. In the latter case, the system is decoupled into simpler subsystems to facilitate the control process (a similar approach is commonly followed in quadrotor control). It is important to note that iSAC is able to successfully control the system using the full dynamics and without relying on separate controllers, unlike, e.g. in \cite{wu2008gain}. Finally, other techniques that have been applied on VTOL aircrafts and unmanned aerial vehicles (UAVs) in general include sliding mode control and backstepping as explained in the following example \cite{bouabdallah2005backstepping}.

\begin{table*}[t!]
\centering
\begin{threeparttable}
\renewcommand{\arraystretch}{1.3}
\caption{{i}SAC parameter values used in simulations}
\label{sac_params}
\centering
\begin{tabular}{c c c c c c c c c}
\hline\hline
Example  & \ref{cartstep}  & \ref{carstep} & \ref{cartrack} &\ref{pvtol_gust} &\ref{quad_flip} &\ref{quadtraj} &\ref{trace_sim}\\
\hline
$\alpha_d$   			   & $-15\, J\big(x_{i}^{\text{\emph{def}}}(\cdot)\big)$     		& \multicolumn{2}{c}{$-100\, J\big(x_{i}^{\text{\emph{def}}}(\cdot)\big)$} 	& $-10\, J\big(x_{i}^{\text{\emph{def}}}(\cdot)\big)$		& \multicolumn{2}{c}{$-5000$}	&$-5000$\\
$T$									   & $1.2$s			& $1.2$s			& $0.35$s		& $3$s			& $3$s		& $2$s & $15$s\\
$t_s$								   &  $0.01$s			&$0.02$s			&$0.03$s		&$0.02$s			&$0.02$s		&$0.05$s &$0.05$s\\
\multirow{2}{*}{$[u_{\text{\emph{min}}}, u_{\text{\emph{max}}}]$} 						   &\multirow{2}{*}{$\,\,\,\,\,\,\,\,\,\,\,\,\,\,\,\,[-20, 20]\,\text{m}/\text{s}^2$}			& \multicolumn{2}{c}{$[-10, 10]\,\text{m}/\text{s}^2$}		& $[0, 5]\,\text{N}$		& \multicolumn{2}{c}{$[0, 12]\,\text{rad}^2/\text{s}^2$}			 
& $[-0.05, 0.05]\,\text{N m}$	 \\
		&		& \multicolumn{2}{c}{$[-4, 4]\,\text{rad}/\text{s}$}		& $[-100, 100]\,\text{N m}$		&\multicolumn{2}{c}{for all inputs}	& for all inputs\\

\hline \hline
\end{tabular}
\end{threeparttable}
\end{table*}

\subsection{Quadrotor}
For this example, a quaternion-based model is used because it leads to polynomial, singularity-free dynamical equations for the quadrotor. The model can be found in \cite{ARCH15:Benchmark_Quadrotor_Attitude_Control} and the model parameters we used are the same as in \cite{Fan-RSS-16}. The 13-dimensional state vector consists of the position and velocity of the center of mass, the angular orientation with respect to the inertial frame expressed in quaternions and the angular velocity expressed in the body frame, i.e., $[x_q, y_q, z_q, \dot{x}_q, \dot{y}_q, \dot{z}_q, q_0, q_1, q_2, q_3, p, q, r]$. The control inputs are the squared angular velocities of the rotors, which are converted to upward thrust for each rotor when multiplied by an appropriate constant. \added{For both examples that follow, nominal control $u^{\text{\emph{nom}}}$ for each input was set to a constant value such that the net upward thrust is approximately equal to gravity force when the quadrotor is parallel to the ground.}

\subsubsection{Flip}\hfill
\label{quad_flip}

To perform a flip, the desired quaternion vector was a flip trajectory with the respective $Q$ weights set equal to $1500$. After the flip, the desired position for the center of mass of the quadrotor was \mbox{$x^d=[1, 1, 1]$} with the respective $Q$ weights set equal to $3$. Finally, \mbox{$R = Diag[0.1\,, 0.1\,, 0.1\,, 0.1]$}. Figures~\ref{fig:results} and ~\ref{fig:costs} show the resulting trajectories and cost.

\subsubsection{Infeasible three dimensional figure eight tracking}\hfill
\label{quadtraj}

In this example, the quadrotor starts in an upside-down position and is set to track a three-dimensional figure eight (respective $Q$ weights set equal to $10$) while keeping the quaternion states at
\mbox{$\big(q_0^d(t), q_1^d(t), q_2^d(t), q_3^d(t)\big)=[1, 0, 0, 0]$} (respective $Q$ weights set equal to $1000$). Thus, the quadrotor is requested to track an aggressive 3-dimensional maneuver without changing its angular orientation, which is infeasible. The weight \mbox{$R = Diag[0.1\,, 0.1\,, 0.1\,, 0.1]$}. As seen in Fig.~\ref{fig:results}, at the beginning of the simulation the quadrotor reverts its angular orientation and then it starts tracking the desired trajectory keeping the error small. This behavior is automatically generated by our method. The corresponding cost is in Fig.~\ref{fig:costs}.
 
Many different approaches have been followed for control of quadrotors over the years. Linear control methods like LQ and PID synthesis \cite{pounds2010modelling,bouabdallah2004pid} are popular for their simplicity. They utilize decoupled or simplified dynamics and also exploit differential flatness \cite{mellinger2011minimum} in order to track feasible trajectories. For example, in \cite{fresk2013full} a quadrotor flip is implemented using a dedicated attitude controller. In contrast iSAC controls the full nonlinear dynamics of the system, and has good performance even with infeasible trajectories as shown in the simulation examples. No additional steps are required to apply our method to this system, unlike, e.g., feedback linearization \cite{lee2009feedback} or sliding mode control \cite{bouabdallah2005backstepping}. The simplicity and versatility of iSAC is one of the main points that this section demonstrates; the same exact approach is being applied to a variety of challenging systems in real time, by replacing only the dynamic model that is being controlled.

\subsection{\added{Incorporating State Constraints -- TRACE Simulation}}
\label{trace_sim}
\added{In 2010, the first in-orbit, time-optimal maneuver was carried out, onboard the NASA's TRACE spacecraft \cite{karpenko2011flight}. The TRACE is a three-axis stabilized, zero-momentum, system that employs a set of four reaction wheels for primary attitude control. The in-orbit maneuvering strategy was calculated by pseudospectral (PS) optimal control theory \cite{fahroo2008pseudospectral,bedrossian2009zero}.}

\added{For this example, the dynamics and parameters of the system were taken from \cite{karpenko2011flight}. The inputs are the torques applied to the reaction wheels. The 11-dimensional state vector consists of the angular orientation expressed as a quaternion vector, the vector of angular body rates and the reaction wheel rates, i.e., $[q_0,q_1,q_2,q_3,\omega_1,\omega_2,\omega_3,\Omega_1,\Omega_2,\Omega_3,\Omega_4]$. The control objective was a rest-to-rest $100^\circ$ time-optimal reorientation about the spacecraft's z-axis. The reaction wheels in the initial configuration are moving at constant rates of $20$~rad/s to provide the necessary energy for the desired maneuver.}

\added{To introduce saturation constraints on the state of the form \mbox{$-x_{\text{\emph{sat}}}<x<x_{\text{\emph{sat}}}$} we included bounded, differentiable penalty functions in the cost as }
\begin{equation}
B(x) = \frac{\bar{Q}}{1+e^{\pm a(x\pm x_{\text{\emph{sat}}})}}
\end{equation}
\added{where $\bar{Q}$ and $a$ are parameters to be selected. These can be directly integrated into the running cost $l$ without violating any of the stated assumptions. For this simulation the only nonzero $Q$ weights corresponded to the quaternion vector and were set equal to $1500$ with $R = Diag[0.1\,, 0.1\,, 0.1\,, 0.1]$, thus prioritizing the maneuvering time. The penalty functions were applied at the body rates $\omega$ and the reaction wheel rates $\Omega$ according to \cite{karpenko2011flight} as $|\omega_i|<0.5$ degrees/s, $i=1,2,3$ and $|\Omega_j|<100$ rad/s, $j=1,2,3,4$. Also, $\bar{Q}=100$ and $a=80$. Finally, nominal control $u^{\text{\emph{nom}}}$ was set to zero for all inputs. The remaining iSAC parameters are given in Table~\ref{sac_params}.}

\added{The iSAC simulation results as well as the real flight results that correspond to the PS method (see \cite{karpenko2012first}) are shown in Fig.~\ref{fig:results}. First, notice that both methods successfully complete the maneuver while satisfying the constraints on the body rates and the wheel rates throughout the simulation. Despite the fact that the PS method directly optimizes the maneuver time in the cost function (we indirectly do so by weighing the state error), it took approximately 170 seconds for iSAC to complete the maneuver, i.e., 10 seconds less than the PS method (181 seconds). Also, the target quaternion vector in the PS method was updated every 1 second, but no relevant discussion was provided for the execution time in \cite{karpenko2012first}. Our simulation with iSAC took less than 40 seconds for the 4-minute trajectory shown in Fig.~\ref{fig:results}, and computations were efficient enough to be executed every $t_s=0.05$s.}

\added{Additionaly, note that both methods follow an off-eigenaxis rotation, i.e., a rotation about a variable axis. The explanation for this lies in the fact that the body masses were not symmetrically distributed on the spacecraft, and as a result, the shortest-time maneuver did not correspond to the shortest angle path. Nevertheless, $\phi$ and $\theta$ in iSAC deviate much less from zero and the same is true for the respective angular velocities $\omega_1$ and $\omega_2$. Finally, the wheel rates $\Omega$ are smaller for almost all time, indicating that iSAC uses less control authority than the PS method over the time horizon.}

\subsection{Discussion and Limitations}
\label{sec:limitations}
In general, iSAC performs well in the variety of scenarios presented. When the task is feasible, trajectories are asymptotically stable and the cost is generally decreasing from time step to time step as a result of applying condition \eqref{new_improve}. Nevertheless, near the desired equilibrium, the behavior of the cost sometimes deviates from what one would expect, even when the control task is feasible. As an example, in the rightmost panel of Fig.~\ref{fig:costs} the cost is not continuously decreasing near the equilibrium, although the system is effectively stabilized when the unexpected behavior occurs (the cost value is less than 0.01). There are three possible explanations for this. First, in order to speed up the solution process, the line search (Section~\ref{howlong}) in all examples presented in this paper was stopped after 10 iterations. While for the majority of cases 10 iterations were sufficient to satisfy the sufficient descent condition \eqref{new_improve}, on some occasions more iterations were necessary. As a result, a small increase was observed in the cost in these time steps. Another factor to consider is numerical tolerance. From the simplified model \eqref{djdlam_first_order}, depending on the value of the mode insertion gradient, the required $\lambda$ value that achieves the sufficient decrease descent could be smaller than the minimum tolerance of numerical integrators. This issue appears often near the equilibrium where systems are more sensitive to the duration of an action. Finally, from Proposition~\ref{exists pair}, it is possible that the selected sampling time $t_s$ was not sufficiently small to (recursively) satisfy \eqref{final_isac_improv}. Since the open-loop problem in iSAC can be solved quickly and efficiently, a strategy for selecting $t_s$ is to start with small values and gradually increase $t_s$ until \eqref{final_isac_improv} is violated.

%% file: conclusion.tex
\label{conclusion}
In this paper we presented iSAC, a model-predictive approach for control of nonlinear systems. As the time horizon progresses, our method sequences together optimal actions and synthesizes piecewise continuous control laws. Some key characteristics of iSAC include: a) analytic solution to the open-loop problem---there is no need to rely on nonlinear programming solvers, b) iterative update of the open-loop solution and c) use of continuous dynamics while incorporating control constraints without additional overhead. Due to these points, iSAC leads to computationally efficient solutions. To establish closed-loop stability, we applied a contractive constraint on the cost. Compared to methods relying on terminal region constraints, the contractive constraint alleviates the need to calculate a terminal region. We also investigated different control scenarios ranging from feasible and infeasible trajectory tracking to set point stabilization with or without external disturbances. Finally, we presented simulation examples using a variety of challenging systems to demonstrate the applicability and flexibility of our method.

%% file: appendix.tex
\label{appendix}
\subsection{Proof of Proposition~\ref{negative_djdl}}
Substituting \eqref{uopt} into the mode insertion gradient formula given in \eqref{Ju} we get (we omit superscripts for brevity)
\begin{align}
\frac{dJ}{d \lambda}(t)&=\Gamma(t)\, \big(\Gamma(t)^T\Gamma(t) + R^T\big)^{-1} \, \Gamma(t)^T \, \alpha_d \notag \\
&=\alpha_d \, ||\Gamma(t)||^2_{(\Gamma(t)^T\Gamma(t) + R^T)^{-1}}<0 \text{,}
\end{align}
where $\Gamma(t)^T = h\big(t, x(t)\big)^T\rho(t)$. The term \mbox{$\Gamma(t)^T\Gamma(t)$} produces a positive semi-definite matrix, and adding \mbox{$R>0$} yields a positive definite matrix. Because the inverse of a positive definite matrix is positive definite, the quadratic norm \mbox{$||\Gamma(t)||^2_{(\Gamma(t)^T\Gamma(t) + R^T)^{-1}}$} is positive for \mbox{$\Gamma(t)^T \neq 0$} (Assumption~\ref{actionability}). Therefore, if \mbox{$\alpha_d<0$}, \mbox{$\frac{dJ}{d \lambda}(t)<0$}.
\hfill\IEEEQEDclosed
\subsection{Proof of Proposition~\ref{exists pair}}
\label{param_exist_proof}
From Proposition~\ref{negative_djdl} and \eqref{djdlam_first_order}, we know that \mbox{$J\big( x_i^*(\cdot)\big) -  J\big(x_{i}^{\text{\emph{def}}}(\cdot)\big) = \Delta J <0$} for each $\lambda$ in a neighborhood around \mbox{$\lambda \to 0^+$} if $\alpha_d <0$. \deleted{For the special case of  $t_s \to 0$, it follows directly from Proposition~\ref{negative_djdl} that~\eqref{final_isac_improv} is always feasible since $C\equiv 0$.}
By continuity assumptions for $m$ and $l$, $C$ in \eqref{final_isac_improv} is continuous with respect to $t_s$ since $t_s$ is part of the integral limits. Thus, it follows that there exists a sufficiently small $t_s$ such that $\Delta J<C$. \hfill\IEEEQEDclosed

\subsection{Proof of Proposition~\ref{existence_of_solution}}
\label{proposition_proof}
The open-loop problem $\mathcal{B}$ is solved by following the same four sequential steps as in the open-loop problem $\mathcal{P}$ in Section~\ref{prelims}. The only difference is that in the solution process of $\mathcal{B}$, the superscripts \emph{nom} are replaced by \emph{def}. To show that $u_i^*(t)$ exists, we will show that each of the four solution steps has a solution: 

\emph{1)} The first step in Section~\ref{prediction} involves calculating the solutions to \eqref{f} and \eqref{rhodot}. Since the default control is in general discontinuous, if solutions are interpreted as sample and hold (CLSS) solutions (see \cite{clarke1997asymptotic}), existence follows directly from Assumptions~\ref{f_assum}, \ref{ml_assum}.

\emph{2)} In the second step our method in Section~\ref{computeoptimal} calculates the optimal action schedule $u_s^*$ by minimizing \eqref{Ju}. Because \eqref{Ju} is convex with a continuous first variation from Assumptions 1-\ref{ml_assum}, solutions \eqref{uopt} exist and are unique, which is also both necessary and sufficient for global optimality of \eqref{Ju}. 

\emph{3)} In the third step in Section~\ref{whentoact} the process selects $\tau_A$ and $u_A$ by minimizing \eqref{Jt}. Because $\tau_A \in [t_i,t_i+T]$ and \eqref{Jt} is in general piecewise continuous (and thus bounded), a solution to this one-dimensional problem exists. 

\emph{4)} Finally, from Proposition~\ref{exists pair}, the backtracking process in Section~\ref{howlong} is guaranteed to find a duration that satisfies condition~\eqref{final_isac_improv} for sufficiently small $t_s$. 

Since all four subproblems have solutions, the open-loop solution $u_i^*(t)$ exists. \hfill\IEEEQEDclosed

\subsection{Proof of Theorem~\ref{theorem1}}
\label{theorem_proof}
The proof has two parts; Lemma~\ref{bounded_M} shows that the integral $\int_{t_0}^{t} M\big(x(s)\big)\,ds$ is bounded for $t\to \infty$ (see Assumption~\ref{ml_assum}). The latter is used in conjunction with a well-known lemma found, e.g., in \cite{michalska1994nonlinear,barbalat1959systemes}, to prove asymptotic convergence in the second part of the proof.

First, define 
\begin{gather}
\label{mpc_cost}
V\big(\alpha, \beta, x(\cdot)\big) = \int_{\alpha}^{\beta} l\big(s,x(s)\big) \,ds + m\big(\beta,x(\beta)\big) \text{.}
\end{gather}
Consider the horizon interval $[t_{i},t_{i}+T]$. Let $u_i^*(t)$ be the solution to the open-loop problem \mbox{$\mathcal{B}(t_i, x_{i})$} and $x_i^*(t)$ the corresponding state trajectory. Clearly, \mbox{$V\big(t_i, t_i+T,x_i^*(\cdot)\big)=J\big(x_i^*(\cdot)\big)$}. Then, for \mbox{$t\in [t_{i},t_{i}+T]$}
\begin{align}
\label{eq:relation}
V\big(t, \,&t_i+T, x_i^*(\cdot)\big) \notag\\
&= V\big(t_i, t_i+T, x_i^*(\cdot)\big) - \int_{t_i}^{t} l\big(s,x_i^*(s)\big) \,ds \text{.}
\end{align}

Based on the above, we now present the following lemma.
\begin{Lemma}
\label{bounded_M}
For small $t_s$, all \mbox{$t\in [t_{i},t_{i}+T]$} and all $i\in \mathbb{N}$
\begin{align}
\label{eq:lemma1}
\int_{t_0}^{t} M\big(\overline{x}(s)\big)ds \leq V\big(t_{0},\,&t_{0}+T,x_{0}^*(\cdot)\big) \notag\\
&-V\big(t,t_i+T,x_{i}^*(\cdot)\big)\text{,}
\end{align}
with
\begin{align}
\overline{x}(t) = \begin{cases} 
      \overline{x}_{cl}(t) & \text{for } t<t_i \\
      x_i^*(t) & \text{else}
   \end{cases} \notag \text{.}
\end{align}

\end{Lemma}
\begin{IEEEproof}
From \eqref{new_improve} and Assumption~\ref{ml_assum} we get that
\begin{align}
\label{intermlem1}
V\big(t_i,t_i+T,x_i^*(\cdot)\big)&-V\big(t_{i-1},t_{i-1}+T,x_{i-1}^*(\cdot)\big) \notag \\
&\leq - \int_{t_{i-1}}^{t_{i}} M\big(x_{i-1}^*(s)\big)ds
\end{align}
holds in \mbox{$[t_{i},t_{i}+T]$} for sufficiently small $t_s$ (Proposition~\ref{exists pair}). Using the corresponding inequalities from the previous time steps until \mbox{$[t_{0},t_{0}+T]$} and the fact that the open-loop solution $u_i^*(\cdot)$ is only applied in \mbox{$[t_{i},t_{i+1}]$} we can write
\begin{align}
\label{intermlem2}
\int_{t_{0}}^{t_i} M\big(\overline{x}_{cl}(t)\big)ds \leq V\big(t_{0},t_{0}&+T,x_{0}^*(\cdot)\big) \notag \\
 &- V\big(t_i,t_i+T,x_i^*(\cdot)\big)\text{.}
\end{align}
Also, from \eqref{eq:relation} we have
\begin{align}
\label{eq:relation2}
 \int_{t_i}^{t} M\big(x_i^*(s)\big) \,ds \leq V\big(t_i, t_i+&T, x_i^*(\cdot)\big) \notag\\
 & - V\big(t, t_i+T, x_i^*(\cdot)\big) \text{.}
\end{align}
Adding \eqref{intermlem2} and \eqref{eq:relation2} leads to \eqref{eq:lemma1} and Lemma~\ref{bounded_M} is proved.
\end{IEEEproof}

From Lemma~\ref{bounded_M}, because $V\big(t,t_i+T,x_{i}^*(\cdot)\big)\geq 0$ and $M$ is positive definite, we can deduce that $\int_{t_0}^{t} M\big(x(s)\big)\,ds$ is bounded for \mbox{$t\to \infty$}. We also have that $x_i^*(\cdot)$, and thus $\overline{x}_{cl}(\cdot)$, are bounded and from the properties of $f$, $\dot{x}_i^*(\cdot)$ and $\dot{\overline{x}}_{cl}(\cdot)$ are bounded as well. These facts combine with the following well-known lemma to prove asymptotic convergence.
\begin{Lemma}
\label{mylemma_hybrid}
Let \mbox{$x:\mathbb{R^+}\to\mathcal{X}$} be an absolutely continuous function and \mbox{$M:\mathcal{X}\to\mathbb{R}^+$} be a continuous, positive definite function ($0 \in \mathcal{X}$). If
\begin{align}
&||x(\cdot)||_{L^\infty (\mathbb{R}^+)} < \infty \text{,} \notag \\
&||\dot{x}(\cdot)||_{L^\infty (\mathbb{R}^+)} < \infty \text{, and} \notag \\
&\lim_{T \to \infty} \int_{0}^{T} M\big(x(t)\big)\,dt < \infty \notag 
\phantom{\hspace{5cm}}
\end{align}
then \mbox{$x(t)\to 0$ as $t \to \infty$}.
\end{Lemma}
\begin{IEEEproof}
The proof can be found, e.g., in \cite{michalska1994nonlinear,barbalat1959systemes}.
\end{IEEEproof}

Theorem~\ref{theorem1} is proved.\hfill\IEEEQEDclosed

\subsection{Proof of Theorem~\ref{cost_lyap}}
\label{theorem_lyap_proof}
We will show that, for all $i$, there exist positive constants $a$, $b$, $c$, such that\\
1. $a ||x_i^*||^2 \leq J\big(t_i, x_i^*(\cdot)\big) \leq b ||x_i^*||^2$ \\
2. $J\big(t_i, x_i^*(\cdot)\big)-J\big(t_{i-1}, x_{i-1}^*(\cdot)\big) \leq -c ||x_{i-1}^*||^2$.

Using Assumption~\ref{bounded_traj}, we can find a constant $d>0$ such that 
\begin{equation}
\label{bound_trans}
||x_i^*(t)||\leq d ||x_i^*|| \text{, } \forall t \in [t_i, t_i+T], \,\,i \in \mathbb{Z}^+
\end{equation}
with $x_i^* = x_i^*(t_i)$. Since $u$ is constrained, this is always true except for systems with finite escape times which are already ruled out from Assumption~\ref{bounded_traj}.

$\bullet$ \emph{Upper bound on $J\big(t_i, x_i^*(\cdot)\big)$}: From \eqref{quadr_J_lyap} and \eqref{bound_trans} we have
\begin{align}
\label{upper_bound_lyap}
J\big(t_i, &x_i^*(\cdot)\big) \leq \frac{1}{2} T \lambda_{\max}(Q) d^2 ||x_i^*||^2 +  \frac{1}{2}  \lambda_{\max}(P_1) d^2 ||x_i^*||^2 \notag \\
&= \frac{d^2}{2} \big(T \lambda_{\max}(Q) + \lambda_{\max}(P_1)\big) ||x_i^*||^2 = b ||x_i^*||^2 \text{.}
\end{align}

$\bullet$ \emph{Lower bound on $J\big(t_i, x_i^*(\cdot)\big)$}:
Using \eqref{bound_trans}, the reverse triangle inequality and \mbox{$||f|| \leq \xi ||x||$} we have
\begin{align}
||x_i^*(t)|| &\geq ||x_i^*|| - \int_{t_i}^{t} ||f|| \,d\tau \notag \\
&\geq ||x_i^*|| - \int_{t_i}^{t} \xi||x_i^*(\tau)|| \,d\tau \notag \\
&\geq [1-\xi d (t-t_i)]||x_i^*|| \notag \text{.}
\end{align}
It then follows, for example, that
\begin{align}
\label{cases_lyap_proof}
||x_i^*(t)|| &\geq \frac{||x_i^*||}{2} \text{ for } t\in \left[t_i, t_i + \frac{1}{2 \xi d}\right] \text{.}
\end{align}
Thus, we have two cases to consider:\\
1) $t_i+T \leq t_i + \frac{1}{2 \xi d}$, or $T \leq \frac{1}{2 \xi d}$.\\
In this case,
\begin{align}
J\big(t_i, x_i^*(\cdot)\big) \geq \frac{1}{2} \int_{t_i}^{t_i+T} \lambda_{\min}(Q)||x_i^*(t)||^2 \,dt \notag \\
 + \frac{1}{2} \lambda_{\min}(P_1)||x_i^*(t_i+T)||^2 \notag
\end{align}
or
\begin{align}
\label{lower_bound_lyap1}
J\big(t_i, x_i^*(\cdot)\big) \geq \frac{1}{8} T \lambda_{\min}(Q)||x_i^*||^2 \text{.}
\end{align}
2) $t_i+T \geq t_i + \frac{1}{2 \xi d}$, or $T \geq \frac{1}{2 \xi d}$.\\
In this case
\begin{align}
J\big(t_i, x_i^*(\cdot)\big) \geq \frac{1}{2} \int_{t_i}^{t_i+\frac{1}{2 \xi d}}  \lambda_{\min}(Q)||x_i^*(t)||^2 \,dt \notag \\
 + \frac{1}{2} \lambda_{\min}(P_1)||x_i^*(t_i+T)||^2 \notag
\end{align}
or
\begin{align}
\label{lower_bound_lyap2}
J\big(t_i, x_i^*(\cdot)\big) \geq \frac{1}{16 \xi d} \lambda_{\min}(Q)||x_i^*||^2 \text{.}
\end{align}
Thus, it follows from \eqref{lower_bound_lyap1}, \eqref{lower_bound_lyap2} that 
\begin{align}
\label{lower_bound_lyap}
J\big(t_i, x_i^*(\cdot)\big) & \geq \min \left\{\frac{1}{8} T \lambda_{\min}(Q), \frac{1}{16 \xi d} \lambda_{\min}(Q) \right\} ||x_i^*||^2 \notag \\
&= a  ||x_i^*||^2 \text{.}
\end{align}
$\bullet$ \emph{Upper bound on $J\big(t_i, x_i^*(\cdot)\big)-J\big(t_{i-1}, x_{i-1}^*(\cdot)\big)$}: From \eqref{new_improve} and \eqref{cases_lyap_proof} we have the following two cases to investigate.\\
1) $t_i \leq t_{i-1} + \frac{1}{2 \xi d}$, or $t_s \leq \frac{1}{2 \xi d}$.\\
In this case,
\begin{align}
\label{cost_bound1}
&J\big(t_i,x_i^*(\cdot)\big) - J\big(t_{i-1},x_{i-1}^*(\cdot)\big) \\
&\leq-\int_{t_{i-1}}^{t_{i}} \lambda_{\min}(Q) ||x_{i-1}^*(t)||^2 \,dt \leq -\frac{1}{4} t_s \lambda_{\min}(Q) ||x_{i-1}^*||^2 \notag \text{.}
\end{align} 
2) $t_i \geq t_{i-1} + \frac{1}{2 \xi d}$, or $t_s \geq \frac{1}{2 \xi d}$.\\
In this case 
\begin{align}
\label{cost_bound2}
J\big(t_i,x_i^*(\cdot)\big) - &J\big(t_{i-1},x_{i-1}^*(\cdot)\big) \notag\\
&\leq- \int_{t_{i-1}}^{t_{i-1}+\frac{1}{2 \xi d}}  \lambda_{\min}(Q) ||x_{i-1}^*(t)||^2 \,dt \notag \\
&\leq -\frac{1}{8 \xi d} \lambda_{\min}(Q) ||x_{i-1}^*||^2  \text{.}
\end{align} 
Thus, it follows from \eqref{cost_bound1}, \eqref{cost_bound2} that 
\begin{align}
\label{final_lyap_decrease}
&J\big(t_i, x_i^*(\cdot)\big) - J\big(t_{i-1},x_{i-1}^*(\cdot)\big) \\
&\leq -\min \left\{\frac{1}{4} t_s \lambda_{\min}(Q), \frac{1}{8 \xi d} \lambda_{\min}(Q) \right\} ||x_{i-1}^*||^2 = -c ||x_{i-1}^*||^2 \notag \text{.}
\end{align}
Theorem~\ref{cost_lyap} is proved.\hfill\IEEEQEDclosed

\subsection{Proof of Theorem~\ref{theorem_dist}}
The difference between the dynamics of the plant and the model used by iSAC for the control computation at each time step $i$ and $t \in [t_i,t_i+T]$ is given by
\begin{align}
\label{temp_diff}
\dot{x}_i^p(t)&-\dot{x}_i^*(t) 
&= f(t,x^p,u_i^*) - f(t,x_i^*,u_i^*) + \eta_i(t)    \text{.}
\end{align}
The states of this model are updated using feedback at every $t_i$, so $x_i^*(t_i)=x^p(t_i)$ for all $i$. Then, we can integrate \eqref{temp_diff} to obtain
\begin{align}
&x_i^p(t)-x_i^*(t) \notag\\
&= \int_{t_{i}}^{t} [f\big(s,x^p(s),u_i^*(s)\big) - f\big(s,x_i^*(s),u_i^*(s)\big)] + \eta_i(s) \,ds  \notag \text{.}
\end{align}
Therefore, using Assumption~\ref{bounded_eta}, we can write
\begin{align}
\label{traj_disc}
&||x_i^p(t)-x_i^*(t)|| \notag \\
&\leq \int_{t_{i}}^{t} ||f\big(s,x^p(s),u_i^*(s)\big) - f\big(s,x_i^*(s),u_i^*(s)\big)|| \,ds \notag \\
& \,\,\,\,\,\,\,\,\,\,\,\,\,\,\,\,\,\,+ \delta_i T := \Delta_i(t) \text{.}
\end{align}
The terminal cost $m$ and the running cost $l$ are Lipschitz continuous from Assumption~\ref{ml_assum}, so we can use this relationship to find the expected cost discrepancy:
\begin{align}
\label{cost_disc}
||J\big(&x_i^p(t)\big)-J\big(x_i^*(t)\big)|| \notag \\
&\leq \int_{t_{i}}^{t_i+T} ||l\big(s,x_i^p(s)\big) - l\big(s,x_i^*(s)\big)|| \,ds \notag \\
&\,\,\,\,+ ||m\big(t_i+T,x_i^p(t_i+T)\big)-m\big(t_i+T,x_i^*(t_i+T)\big)|| \notag \\
&\leq \int_{t_{i}}^{t_i+T} L_1 \Delta_i(t_i+T) \,ds + L_2 \Delta_i(t_i+T)  \notag \\
&\leq \Delta_i(t_i+T) (L_1 T + L_2)\text{.}
\end{align}
with Lipschitz constants $L_1$ and $L_2$. Using \eqref{cost_disc} one can show that the upper bound on the contractive constraint \eqref{new_improve} will inevitably be higher due to the effect of the disturbance. Specifically, from \eqref{new_improve} and \eqref{cost_disc}
\begin{align}
\label{contr_dist}
J\big(x_i^*&(\cdot)\big) - J\big(x_{i-1}^*(\cdot)\big) \leq-\int_{t_{i-1}}^{t_{i}} l\big(t, x_{i-1}^*(t)\big) \,dt \notag \\
& + (L_1 T + L_2)[\Delta_i(t_i+T)+\Delta_{i-1}(t_{i-1}+T)]\text{.} 
\end{align} 
This practically means that the cost is allowed to increase for as long as the disturbance is active.

The following Lemma is the last step before proving the theorem.
\begin{Lemma}
\label{delta_to_zero}
As $t \to \infty$, $\Delta_i(t) \to 0$ in \eqref{traj_disc}.
\end{Lemma}
\begin{IEEEproof}
From Assumption~\ref{bounded_eta}, \mbox{$\delta_i \to 0$} as \mbox{$i\to \infty$}. Also, using the fact that \mbox{$x_i^*(t_i)=x^p(t_i)$} and Assumption~\ref{f_assum}, \mbox{$x_i^p(t) \to x_i^*(t)$} as $i \to \infty$.
\end{IEEEproof}
We can now follow the steps in the proof of Theorem~\ref{theorem1} to show asymptotic stability under the disturbance $\eta(t)$. First, we can use Assumption~\ref{ml_assum}, \eqref{contr_dist} and Lemma~\ref{delta_to_zero} to show that  $\int_{t_0}^{t} M\big(x(s)\big)\,ds$ is bounded for \mbox{$t\to \infty$} by following the same procedure as in Lemma~\ref{bounded_M}. Asymptotic stability then follows directly from Lemma~\ref{mylemma_hybrid}. The process is left as an exercise to the reader.

Theorem~\ref{theorem_dist} is proved.\hfill\IEEEQEDclosed